\documentclass{article}

\usepackage{PRIMEarxiv}

\usepackage[utf8]{inputenc} 
\usepackage[T1]{fontenc}    
\usepackage{hyperref}       
\usepackage{url}            
\usepackage{booktabs}       
\usepackage{amsfonts}       
\usepackage{nicefrac}       
\usepackage{microtype}      
\usepackage{lipsum}
\usepackage{fancyhdr}       
\usepackage{graphicx}       
\graphicspath{{media/}}     

\usepackage{hyperref}       
\usepackage{url}            
\usepackage{booktabs}       
\usepackage{amsfonts}       
\usepackage{nicefrac}       
\usepackage{microtype}      
\usepackage{lipsum}
\usepackage{fancyhdr}       
\usepackage{graphicx}
\usepackage{authblk}
\usepackage{forest}
\usepackage{caption}
\usepackage{subcaption}
\usepackage{amsmath}
\usepackage{diagbox}
\usepackage{dirtree}
\usepackage{array}
\usepackage[table]{xcolor}

\usepackage{todonotes}

\title{CESNET-TimeSeries26: A Multi-Year Network Traffic Time Series Dataset for Annual Seasonality, Concept Drift, and Long-Term Anomaly Detection
}

\author[1,*]{Josef Koumar}
\author[1]{Karel Hynek}
\author[2]{Kamil Je\v{r}\'{a}bek}
\author[2]{Pavel \v{S}i\v{s}ka}
\affil[1]{Czech Technical University in Prague, Th\'{a}kurova 9, 160 00 Prague 6, Czech Republic}
\affil[2]{CESNET a.l.e., Gener\'{a}la P\'{i}ky 430/26, 160 00 Prague 6, Czech Republic}

\affil[*]{corresponding author: Josef Koumar (josef.koumar@fit.cvut.cz)}

\begin{document}
\maketitle

\begin{abstract}
    Forecast-based models are fundamental to network traffic management and anomaly detection. However, it is essential to evaluate these forecasting models over extended periods of time in order to assess their adaptability to unexpected phenomena and shifts in underlying patterns. This kind of long-term evaluation requires multi-year data to account for yearly seasonality and behavioral drift. Existing public datasets are too short to support such long-horizon forecasting tasks. We introduce CESNET-TimeSeries26, comprising 995 days of continuous traffic time series from the CESNET3 network, together with a single subnet spanning approximately 4.5 years. The dataset captures 278{,}466 IP addresses, 285 institutions, and 566 institutional subnets at three aggregation intervals. The dataset provides a definitive longitudinal benchmark for evaluating forecasting models in both operational management and security monitoring. We demonstrate that annual seasonality is the dominant periodic component. Moreover, we show that a forecasting model can exploit the annual component only when its training window covers at least one complete annual cycle, which no sub-year dataset provides.
\end{abstract}


\section*{Background \& Summary}

    Traffic monitoring is a cornerstone of network management and security operations~\cite{d2019survey}. As traffic encryption became ubiquitous, visibility into network communication decreased, and the research community increasingly turned to machine-learning-based classification of encrypted traffic~\cite{aceto2019mobile,koumar2024nettisa,akbari2022traffic,plny2022decrypto}. However, recent findings demonstrate substantial limits of supervised machine learning approaches used in the networking domain~\cite{jerabek2025simple}. Unsupervised anomaly detection~\cite{10.1007/978-3-642-30507-8_13,8990084,yaacob2010arima,andrysiak2014network,jerabek2025explainable} can reveal malicious activities through behavioral changes in network entities without directly learning malicious patterns.  A large family of unsupervised detectors is built on traffic forecasting, in which an alert is raised when the observed value deviates from the forecast by more than a specified threshold~\cite{yaacob2010arima,andrysiak2014network,jerabek2025explainable}. Therefore, the quality of such detectors is bound by the quality of the underlying forecasting models. Moreover, it is also directly bound by the data available to train and evaluate them.

    The scarcity of long-term, real-world network time series has been repeatedly identified as the crucial obstacle in this area~\cite{ferreira2023forecasting,wu2021current}. Publicly available datasets are predominantly synthetic, which leads to overestimated performance and an ``illusion of progress''~\cite{wu2021current}, while real-world traces such as the MAWI/WIDE captures~\cite{cho2000traffic,mawilab} are limited to 15-minute daily snapshots. We addressed part of this gap with the CESNET-TimeSeries24 dataset~\cite{koumar2025cesnet}, which provided 40 weeks of traffic time series for 275 thousand IP addresses of the CESNET3 network and has since been adopted by the community for benchmarking forecasting and anomaly detection methods. The CESNET-TimeSeries24 became immediately popular in the community with tens of reuses~\cite{chang2026time,liu2025time,koumar2025comparative,feng2026robust,sengendo2026prediction,misa2025sustainability,zhang2026centile,babai2026deep}.

    However, a 40-week collection has a structural limitation: it does not contain a single complete annual cycle. Therefore, the annual seasonal component cannot be fitted, the recurrence of holiday and vacation patterns cannot be distinguished from one-off anomalies, and the multi-year drift of network behavior cannot be quantified at all. In an academic ISP network, the annual rhythm of terms, exam periods, holidays, and vacations is the strongest low-frequency signal in the data. Our analyses confirm that annual seasonality is, in fact, the \emph{dominant} periodic component for the vast majority of entities. Therefore, sub-year datasets are structurally blind to it.

    This manuscript introduces CESNET-TimeSeries26, created by continuous long-term monitoring of the CESNET3 network over 995 days---from 9~October~2023 to 30~June~2026. The collection uses the identical monitoring infrastructure, capturing pipeline, metrics, anonymization, and processing pipeline as CESNET-TimeSeries24. The collection has been permanently closed due to a major change in the monitoring infrastructure, meaning new data points can no longer be added to the time series. Therefore, it is final and constitutes a stable, definitive reference for longitudinal studies. It was created from approximately 190.3 billion IP flows containing 13.6 trillion packets and 13.5 petabytes of transferred data, and provides time series for 278 thousand IP addresses, 285 institutions, and 566 institutional subnets at 10-minute, 1-hour, and 1-day aggregation intervals. In addition, the longitudinal part covers a single institutional subnet for which archived flow-level data reaches back to 2022, yielding approximately 4.5 years of continuous, TCP-only time series.

    The scientific contribution of this descriptor is not that longer data yields incrementally better forecasts. In fact, we show that it does not in the point-forecast sense. The contribution is that a multi-year span makes an entire class of phenomena \emph{measurable on public network data for the first time}. Multi-year public time series do exist in other domains, e.g., road traffic~\cite{liu2023largest}, retail sales~\cite{makridakis2022m5}, and electricity demand or pedestrian counts~\cite{godahewa2021monash}, yet the benchmarks most widely used for long-term forecasting span at most two to four years~\cite{lai2018modeling,zhou2021informer,wu2021autoformer}, and no comparable span has so far been publicly available for network traffic. 
    
\begin{figure*}
        \centering
        \includegraphics[width=\textwidth]{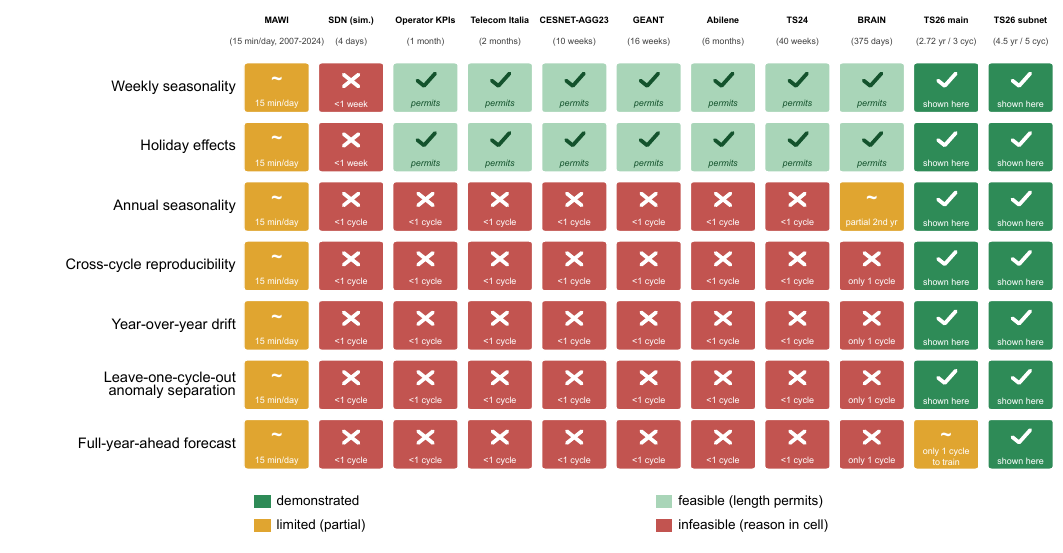}
        \caption{Analyses unlocked by a length of dataset. Columns are eleven network-traffic datasets ordered by span: MAWI~\cite{mawilab}; SDN~\cite{9500331}; Network Operator KPIs~\cite{fondo_ferreiro_2023_8147768}; the Telecom Italia~\cite{barlacchi2015multi}; CESNET-AGG23~\cite{benes2023look}; GEANT~\cite{10.1145/1111322.1111341}; Abilene~\cite{10.1145/781027.781053}; the 40-week CESNET-TimeSeries24~\cite{koumar2025cesnet} (TS24); BRAIN~\cite{orlowski2010sndlib}; CESNET-TimeSeries26~\cite{CESNET_TIMESERIES26_DATASET} main part (TS26 main); and CESNET-TimeSeries26~\cite{CESNET_TIMESERIES26_DATASET} longitudinal subnet part (TS26 subnet). A MAWI archive of 15-minute daily snapshots is included separately: although it spans calendar years (2007--2024), each captured day is only 15 m inutes long, so it never accumulates a single continuous cycle. A full annual cycle is required for annual seasonality. Two cycles for cross-cycle reproducibility, year-over-year drift, and leave-one-cycle-out anomaly separation. Three cycles for a full-year-ahead forecast validated on a held-out complete cycle. Cells marked \emph{demonstrated} are validated in this article. The \emph{feasible} cells are supported by the data length but were not separately run. The \emph{limited} cells denote partial capability. And finally, the \emph{infeasible} cells state the limiting reason.}
        \label{fig:capability_matrix}
    \end{figure*}

    Concretely, our technical validation establishes four findings, each impossible in a sub-year collection: (i)~annual seasonality is the dominant periodic component even stronger than the weekly component for 96.8\% of institutions, 98.2\% of institutional subnets, and 99.8\% of dense IP series; (ii)~the annual component of traffic cannot be exploited by a forecasting model trained on less than one year of data, but it becomes exploitable once the training window spans one complete annual cycle; (iii)~most network entities exhibit statistically significant year-over-year distributional drift, and a slow multi-year structural shift is visible in metrics such as the average TTL; and (iv)~a forecasting model deployed without retraining degrades by approximately 11.3~SMAPE points per quarter, implying a practical retraining interval of about two to three months. The Technical Validation presents the evidence for statements (i) and (iii), and its closing subsection, Relationship to CESNET-TimeSeries24, presents the evidence for statements (ii) and (iv). Table~\ref{fig:capability_matrix} summarizes which analyses each observation length unlocks and locates CESNET-TimeSeries26 relative to sub-year traces and the previous 40-week collection.

\section*{Methods} \label{methods}

    In this section, we describe the methods used to obtain the CESNET-TimeSeries26 dataset. The collection methodology is identical to the one used for CESNET-TimeSeries24~\cite{koumar2025cesnet}. We summarize it here for self-containment and describe in detail the aspects specific to the multi-year collection.

    \subsection*{Ethical statement}

        Since the dataset originates from a production network used by real users, privacy was a fundamental concern throughout our work. We used exclusively automated data processing with immediate anonymization. Therefore, we did not analyze or manually process non-anonymized data at any point, nor did we perform any procedure that would allow us to track users or reveal their identities.

        The publication of the dataset has been approved by the Committee for Ethics in Research of the Scientific Council of the Czech Technical University in Prague under reference number 0000-11/26/51902/EKČVUT. The approval includes a waiver of explicit user consent for publishing the dataset. Moreover, all users of the CESNET3 network agreed to terms and conditions that define a monitoring process for the optimization and improvement of services (including related research) and allow sharing of the data with third parties after anonymization (\url{https://www.cesnet.cz/en/gdpr}).

    \subsection*{Data capture}

\begin{figure*}
            \centering
            \includegraphics[width=13cm]{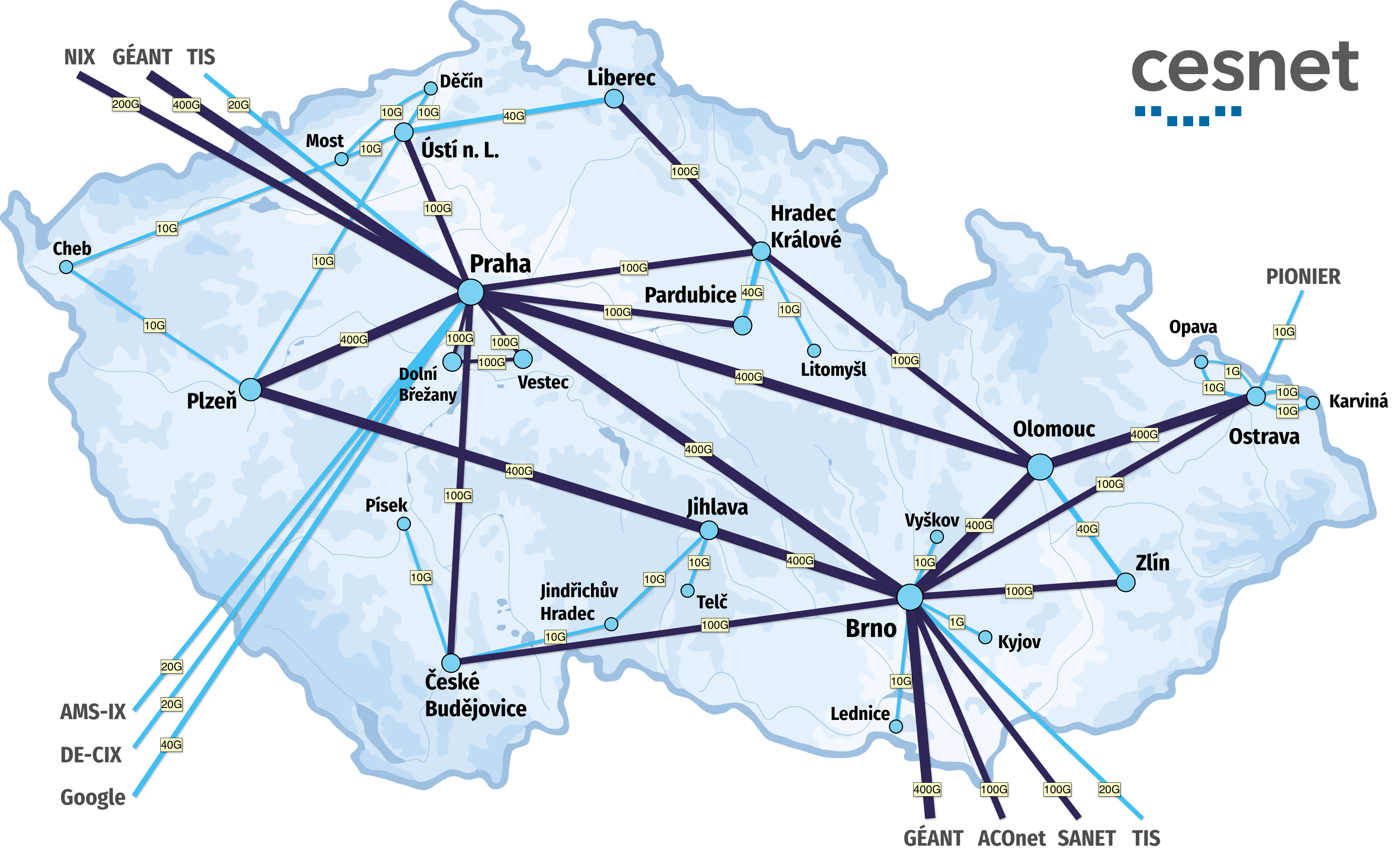}
            \caption{Topology of the CESNET3 network, which interconnects academic institutions in the Czech Republic.}
            \label{fig:topology_cesnet2}
        \end{figure*}

        The network traffic was obtained from the ISP network CESNET3, which provides internet access to public and research institutions in the Czech Republic. The network spans the whole country, as shown in Figure~\ref{fig:topology_cesnet2}, and serves approximately half a million users daily. Because packet-based monitoring is infeasible at the ISP scale, CESNET3 is monitored with a standard IP flow monitoring system located at the network perimeter. All transit lines to peering partners are equipped with flow monitoring probes.

        IP flow monitoring systems aggregate packets into IP flow records. An IP flow record represents communication metadata associated with a single connection and is defined~\cite{rfc7011} as ``a set of IP packets passing an observation point in the network during a certain time interval, such that all packets belonging to a particular flow have a set of common properties''. The flow key is defined by source and destination IP addresses, transport-layer ports, and protocol.

\begin{figure*}
          \centering
          \includegraphics[width=0.7\textwidth]{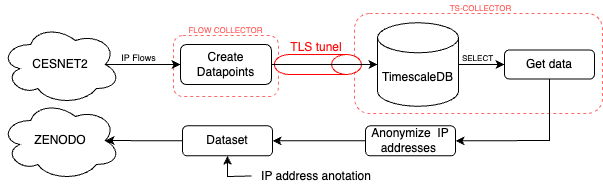}
          \caption{Architecture of dataset collection from the CESNET3 network.}
          \label{fig:dataset_collection}
        \end{figure*}

        The monitoring infrastructure is detailed in Figure~\ref{fig:dataset_collection}. Network TAPs positioned before the edge routers mirror the traffic to monitoring probes, which are servers equipped with high-speed monitoring cards capable of handling 200\,Gb/s. On the servers, the ipfixprobe flow exporter (\url{https://github.com/CESNET/ipfixprobe}) is running with a configuration of an active timeout of 5 minutes and an inactive (idle) timeout of 65 seconds, which is standard practice for flow-based monitoring~\cite{rfc7011}. The exporter collected the following features: start time, end time, number of packets, number of bytes, and Time to Live (TTL). None of the collected information contains application-level payload, which ensures that the privacy of user communication is not compromised. The collected records were transferred over a secure (TLS) IPFIX~\cite{rfc7011} channel to a flow collector running IPFIXcol2 (\url{https://github.com/CESNET/ipfixcol2}).

    \subsection*{Continuity of the measurement infrastructure}

        Because the dataset is intended to enable users to distinguish genuine network phenomena from measurement artifacts, we characterize the continuity of the capture pipeline over the full capture window. Before doing so, we fix the temporal reference frame, since the window spans several civil-time changes.

        The Czech Republic observes Central European Time (CET, UTC$+1$) in winter and Central European Summer Time (CEST, UTC$+2$) in summer, switching on the last Sunday of March and October. Six such transitions fall within the 995-day main window, and nine within the 2022--2026 span of the longitudinal subnet; all nine are listed in Table~\ref{tab:dst}. The dataset's canonical time index is defined on a continuous absolute (UTC) grid, and the Europe/Prague local timestamps reported in this section and in Table~\ref{tab:outages} are derived from it. On this grid, the spring-forward transitions are fully transparent with no window created or lost. However, the fallback transitions repeat within a local hour and would each appear as a spurious one-hour discontinuity if diagnosed by local clock time. Therefore, we identify them by the accompanying $-1$\,h change in UTC offset rather than by the apparent gap length, and treat them as timekeeping artifacts rather than outages.

        \begin{table*}
            \centering
            \caption{\label{tab:dst}Daylight-saving transitions within the CESNET-TimeSeries26 observation window (Europe/Prague). Spring-forward events skip a local hour and are transparent on the absolute (UTC) time grid; fall-back events repeat a local hour and are identified by the $-1$\,h offset change.}
            \rowcolors{3}{gray!10}{white}
            \begin{tabular}{l|l|l|r}
                \textbf{Date} & \textbf{Transition} & \textbf{Local clock change} & \textbf{Offset change} \\
                \hline
                2022-03-27 & spring forward & 02:00 CET  $\rightarrow$ 03:00 CEST & $+1$\,h \\
                2022-10-30 & fall back & 03:00 CEST $\rightarrow$ 02:00 CET  & $-1$\,h \\
                2023-03-26 & spring forward & 02:00 CET  $\rightarrow$ 03:00 CEST & $+1$\,h \\
                2023-10-29 & fall back & 03:00 CEST $\rightarrow$ 02:00 CET  & $-1$\,h \\
                2024-03-31 & spring forward & 02:00 CET  $\rightarrow$ 03:00 CEST & $+1$\,h \\
                2024-10-27 & fall back & 03:00 CEST $\rightarrow$ 02:00 CET  & $-1$\,h \\
                2025-03-30 & spring forward & 02:00 CET  $\rightarrow$ 03:00 CEST & $+1$\,h \\
                2025-10-26 & fall back & 03:00 CEST $\rightarrow$ 02:00 CET  & $-1$\,h \\
                2026-03-29 & spring forward & 02:00 CET  $\rightarrow$ 03:00 CEST & $+1$\,h \\
                \hline
            \end{tabular}
        \end{table*}

        During the multi-year capture, the monitoring infrastructure was occasionally unavailable due to software or hardware maintenance and transient faults. Table~\ref{tab:outages} lists the outages. In total, 13 outages are reported, amounting to approximately 31.5 days of missing observation. Five span more than a day, four last between half a day and a day, and four last one to two hours. Outages 1--6 fall before the start of the main collection and therefore affect only the longitudinal subnet. Outages 7--13 fall within the main window and affect both parts. Shorter gaps and daylight-saving transition artifacts are omitted as negligible. These interruptions are an unavoidable consequence of multi-year operation, where maintenance and faults occur naturally. Users requiring strict continuity can mask these intervals directly from the shared time grid. We do not recommend imputation, particularly for the 15-day and 5-day gaps.

        \begin{table*}
            \centering
            \caption{\label{tab:outages}Monitoring outages in CESNET-TimeSeries26. Times are Europe/Prague local. The Part column states which dataset part the outage affects. Shorter gaps and daylight-saving transition artifacts are omitted. The 2024 probe outage was a partial capacity loss (total never reached zero) and is discussed in the text rather than listed here.}
            \rowcolors{3}{gray!10}{white}
            \begin{tabular}{l|l|l|l|r}
                \textbf{Outage} & \textbf{Part} & \textbf{Last data before} & \textbf{Data resumed} & \textbf{Duration} \\
                \hline
                1  & subnet & 2022-01-11 12:43 & 2022-01-12 12:43 & 23.8 h \\
                2  & subnet & 2022-03-14 22:53 & 2022-03-15 13:33 & 14.5 h \\
                3  & subnet & 2022-05-18 21:53 & 2022-05-19 21:43 & 23.7 h \\
                4  & subnet & 2022-12-11 17:13 & 2022-12-14 10:23 & 2 d 17.0 h \\
                5  & subnet & 2022-12-28 09:53 & 2023-01-02 14:33 & 5 d 4.5 h \\
                6  & subnet & 2023-04-12 01:33 & 2023-04-14 05:43 & 2 d 4.0 h \\
                8  & main, subnet & 2025-10-05 11:57 & 2025-10-05 13:09 & 1.0 h \\
                7  & main, subnet & 2025-10-05 14:13 & 2025-10-06 05:54 & 15.5 h \\
                9  & main, subnet & 2025-10-07 08:17 & 2025-10-07 10:07 & 1.7 h \\
                10 & main, subnet & 2026-03-06 15:07 & 2026-03-08 18:17 & 2 d 3.0 h \\
                11 & main, subnet & 2026-04-24 07:50 & 2026-04-24 09:38 & 1.7 h \\
                12 & main, subnet & 2026-04-29 10:38 & 2026-05-15 07:46 & 15 d 21.0 h \\
                13 & main, subnet & 2026-05-22 11:56 & 2026-05-22 13:16 & 1.2 h \\
                \hline
            \end{tabular}
        \end{table*}

        Moreover, there was a probe outage from 2024-05-21 to 2024-06-04, which was a \emph{partial} capacity loss rather than a complete blackout. The institution's total fell to a median of about 43\,\% of its baseline (minimum 6.4\,GB per 10-minute window) but never reached zero and produced no grid gap. It is consequently absent from the zero-traffic table above and is instead flagged in the known-measurement-artifacts documentation as a reduced-capacity event, so that users do not mistake the partial recovery for a genuine level shift or trend anomaly.

    \subsection*{Time series aggregation}

        The flow collector runs the aggregation module, as depicted in Figure~\ref{fig:dataset_collection}. The aggregation module also removes transient traffic (flows in which neither endpoint belongs to CESNET3) and all single TCP SYN packet connections, i.e., scans. Given the probe placement before routers and firewalls, scans would otherwise dominate the collected flows and introduce significant noise into the time series. Since scans are reliably detectable by simple, dedicated methods~\cite{staniford2002practical,bhuyan2011surveying}, we exclude them. All remaining flows are passed to aggregation.

        \begin{figure*}
          \centering
          \includegraphics[width=\textwidth]{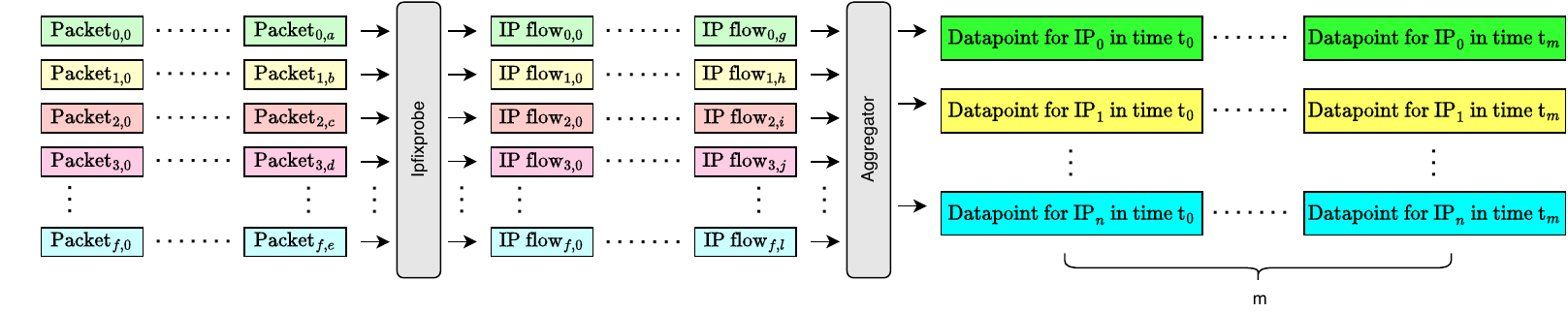}
          \caption{The aggregation process for capturing time series of length $m$ from network traffic. For each packet $Packet_{i,j}$, there exists a $Flow_{i,k}$ to which the packet belongs. It always holds that $a \ge g$ (and similarly $b \ge h$), and in most cases $a$ is much larger than $g$; commonly, a single IP flow contains all packets of one connection ($g = 0$), e.g., for a user visiting a web page. A time series datapoint is likewise a combination of one or more IP flows.}
          \label{fig:traffic_preprocesing}
        \end{figure*}

        We generate evenly spaced time series for each IP address by aggregating IP flow records into data points representing the address's behavior within 10-minute time windows. This process is shown in Figure~\ref{fig:traffic_preprocesing}. The vector of time-series metrics $v_{ip, i}$ describes the IP address $ip$ in the $i$-th window, i.e., between $t_i$ and $t_{i+1}$. Aggregated datapoints are stored in TimeScaleDB (\url{https://www.timescale.com/}), from which the time series are built.

        Datapoints contain the following time-series metrics:
        \begin{itemize}
          \item \textit{\textbf{Simple volumetric metrics:}} the number of IP flows, the number of packets, and the transmitted data size (number of bytes).
          \item \textit{\textbf{Unique volumetric metrics:}} the number of unique destination IP addresses, unique destination Autonomous System Numbers (ASNs), and unique destination transport-layer ports. Aggregation of these metrics is memory-intensive since all unique values must be kept in memory; a server with 41\,GB of RAM was sufficient for 10-minute aggregation at ISP scale.
          \item \textit{\textbf{Ratio metrics:}} the ratio of UDP/TCP packets, the ratio of UDP/TCP transmitted data size, the direction ratio of packets, and the direction ratio of transmitted data size.
          \item \textit{\textbf{Average metrics:}} the average flow duration and the average Time To Live (TTL).
        \end{itemize}

    \subsection*{Anonymization}

        The capture process started on 9~October~2023 and ended on 30~June~2026, i.e., after 995 days of continuous collection, at which point we extracted the time series from the database. On 30 June 2026, the capture infrastructure was decommissioned due to major changes to the monitoring infrastructure that prevented it from continuing to capture long same-origin time series. A script deployed within the database framework automatically assigns a corresponding institution and institutional subnet to each IP address. We do not extract or store IP addresses, institution names, or exact subnets. Instead, randomly assigned database identifiers are used for all three entity types. Without access to the mapping table within the database framework, this step cannot be reversed, and the published time series cannot be linked to a specific IP address, institution, or institutional network.

    \subsection*{Longitudinal subnet collection}

        For one institutional subnet with 255 addresses (subnet with network mask /24,  approximately 100 active), archived TCP flow-level data were available from an earlier deployment of the same monitoring infrastructure, starting in 1~January~2022. We re-aggregated these archived IP flow records into time-series data points using the same filtration and aggregation pipeline, metrics, time windows, and anonymization as for other time series in this dataset. The result spans approximately 4.5 calendar years and is published as a separate part of the dataset. Thus, it contains four complete calendar-year cycles and a partial fifth. Moreover, keyed to the academic year (starting approximately 20~September in Czechia), the span covers three complete annual cycles (2022--23, 2023--24, 2024--25) plus a partial fourth (2025--26), which is what our cross-cycle analyses use.

        The deployment that produces this dataset part exports \emph{only TCP flows}. As a consequence, all volumetric and average metrics of this part (\texttt{n\_flows}, \texttt{n\_packets}, \texttt{n\_bytes}, \texttt{n\_dest\_ip}, \texttt{n\_dest\_asn}, \texttt{n\_dest\_port}, \texttt{avg\_duration}, \texttt{avg\_ttl}) reflect TCP traffic exclusively, and their absolute levels are not comparable with the all-protocol parts of the dataset. The two transport-ratio metrics (\texttt{tcp\_udp\_ratio\_packets} and \texttt{tcp\_udp\_ratio\_bytes}) are constant and equal to~1 over the whole span. They are retained solely for format compatibility with the other parts and carry no information for this part.

        This longitudinal part is self-contained. It covers the 2022--2026 period, including the interval that overlaps with the main dataset. Self-containment duplicates the 2023--2026 data for these addresses (which also appear, as all-protocol series, in the main IP-address and subnet parts) but spares users the need to stitch two sources. The overlapping period is \emph{not} identical between the two views. The main dataset part contains all metrics and is produced from all-protocol traffic. The anonymized identifiers for these addresses in the longitudinal part are identical to those in the main dataset, so users can relate the two views while remaining aware of the TCP-only versus all-protocol difference.

    \subsection*{Data preprocessing}

        Preprocessing consisted of: \textit{i)} time series filtering, \textit{ii)} creation of multiple time aggregations, \textit{iii)} creation of institution time series, \textit{iv)} creation of institutional subnet time series, \textit{v)} creation of sample IP addresses datasets, and \textit{vi)} adding information about weekends and public holidays, school terms and exam periods, and European time changes.

        \paragraph{Filtering} CESNET-TimeSeries24 removed series with more than 99.75\% empty datapoints, resulting in dropping time series with fewer than 100 datapoints. We treat this published rule as a \emph{relative} principle rather than as the absolute constant~100, and recompute it for the 995-day window in the main dataset and the 1651-day window in the longitudinal subnet. With $995\times144 = 143{,}280$ and $1651\times144 = 237{,}744$ theoretical 10-minute windows, the threshold becomes $0.25\%\times143{,}280 \approx 358$ and $0.25\%\times237{,}744 \approx 594$. Thus, we remove series with at most 358 non-empty 10-minute datapoints from the main part and 594 non-empty 10-minute datapoints from the longitudinal subnet part. After filtering, the main dataset contains time series for 278{,}466 IP addresses, 285 institutions, and 566 institution subnets. The longitudinal subnet part contains 135 IP address series.

        \paragraph{Multiple time aggregation} The finest aggregation level groups traffic into 10-minute windows. The window length is bounded by the flow-export process because the time series are built from IP flow records exported with an active timeout of 5 minutes. So a long-lived connection appears as a sequence of records, each carrying up to 5 minutes of traffic. Therefore, windows shorter than the active timeout would distribute the volume of such flows unevenly. Moreover, it would produce spikes and empty intervals that were not actually observed in traffic. Because export boundaries are not aligned with window boundaries, even a 5-minute window would not be enough. Therefore, we set the window to twice the active timeout, which keeps this effect small. Moreover, 10-minute intervals can be too short for longitudinal methods. Therefore, we additionally provide 1-hour and 1-day aggregations derived from the 10-minute data. Where possible, metrics were aggregated by summing values (IP flows, packets, transmitted bytes). For metrics that represent counts of unique values, which cannot be summed without information loss, we provide the sum, mean, and standard deviation, resulting in three derived metrics per original metric. Ratio and average metrics are aggregated using the mean.

        \paragraph{Time series of institutions} Many security events are visible only from the perspective of overall network traffic. Therefore, we use the institution ID exported from the database to group IP addresses by institution. We identify 285 institutions within the CESNET3 network. From which 283 are unchanged from CESNET-TimeSeries24, and 2 new institutions were added during the longer-term capture pipeline.

        \paragraph{Time series of institutional subnets} Institutions often operate multiple networks in different locations or for different purposes, and administrators commonly handle security per subnet. We therefore also group IP addresses by institutional subnets using information from CESNET. We identify 566 institutional subnets within the CESNET3 network. From which 548 are unchanged from CESNET-TimeSeries24, and 18 new institution subnets were added during the longer-term capture pipeline.

        \paragraph{Creation of sample IP addresses datasets} CESNET-TimeSeries24 provided a single uniformly random sample of IP address series, which is usable, for example, for building a method without evaluation on the whole corpus, while still comparing the method with related works. And if the method had promising results on the sample dataset, then the long computation on the whole corpus makes sense. 
        
        We retain a comparable uniform-random baseline subset (\texttt{ip\_addresses\_sample}), which reproduces the original sampling. Moreover, we additionally publish five subsets stratified by percentage of naturally occurring gaps in network traffic time series, so that methods can be evaluated across the full range of gap severity rather than only on the ``average'' series. For each series, we compute a gap ratio score on the 10-minute grid. If the gap ratio is equal to 0, it denotes a fully dense series, and values approaching $1$ denote a mostly absent one. We partition $[0,1]$ into five equidistant bands of width $0.20$ (\texttt{gaps\_20\_00}: $[0.00,0.20]$ near-complete to dense; \texttt{gaps\_40\_20}: $(0.20,0.40]$; \texttt{gaps\_60\_40}: $(0.40,0.60]$; \texttt{gaps\_80\_60}: $(0.60,0.80]$; \texttt{gaps\_100\_80}: $(0.80,1.00]$ very sparse). From each band, we draw $N=1000$ series with a single \texttt{numpy} random generator (seed $42$) consumed in a fixed order over an \texttt{id\_ip}-sorted candidate pool, making the draw independent of filesystem ordering. The candidate pools all exceed $N$ ($7{,}852$, $10{,}869$, $12{,}321$, $19{,}037$, and $228{,}387$ series across the five bands, from densest to sparsest). Thus,  every subset is exactly sized with no out-of-band members. Each sampled series is materialized identically across all three aggregations to keep the subsets comparable.

        \paragraph{Weekends and public holidays} Information about non-working days is crucial for training and evaluating forecasting and anomaly detection models. We include weekend dates and Czech public holidays for the entire collection period, i.e., the years 2022--2026. Because of the multi-year span, every annual holiday (Christmas, Easter, state holidays) occurs at least twice in the main dataset and at least four times in the subnet, enabling us to distinguish recurring calendar effects from one-off anomalies.

        \paragraph{School terms and exam periods} CESNET3 interconnects universities, academic institutions, high schools, student dormitories, and other public institutions in the Czech Republic. Therefore, its traffic closely follows the academic calendar. CESNET-TimeSeries24 could only comment on this correlation, whereas the multi-year span provides the calendar explicitly and observes each phase repeatedly. For the whole collection period, we annotate every date with the prevailing Czech academic-year regime: autumn (winter-semester) teaching term, winter examination period, spring (summer-semester) teaching term, summer examination period, and vacation. Following the official academic calendars of the covered years and keyed to the Czech term start, mostly in the second half of September, however, the start of the winter term is not common for all universities in Czechia. Thus, it can deviate by a week. 
        
        Each annual cycle appears at least twice in the main dataset (academic years 2023/24, 2024/25, and a partial 2025/26) and four times in the longitudinal subnet (2022/23 through 2025/26). This information lets users condition training and evaluation on the operative regime and, because each phase recurs, separate the stable academic structure from one-off events.

        \paragraph{European time changes} The Czech Republic changes its clocks twice a year. On the last Sunday of March, clocks move one hour forward (CET to CEST). On the last Sunday of October, they move one hour back (CEST to CET). The time series in the dataset are not affected by these changes directly. Each series is indexed by a sequence number of its aggregation window, and the windows follow an absolute timeline. The \texttt{times} translation tables map each sequence number to the corresponding wall-clock time. Therefore, no windows are added or removed during a clock change. However, the wall-clock time associated with a given window index shifts by one hour with each change. Because human activity follows local time, the daily and weekly patterns in the data also shift by one hour relative to the window index at each change. Therefore, we provide the exact instances of all clock changes in the 2022--2026 period. Users can align the windows with local time and avoid mistaking this one-hour shift for a real change in a series' behavior.

\section*{Data Records}

    \begin{figure*}[t]
        \centering
        \includegraphics[width=0.7\linewidth]{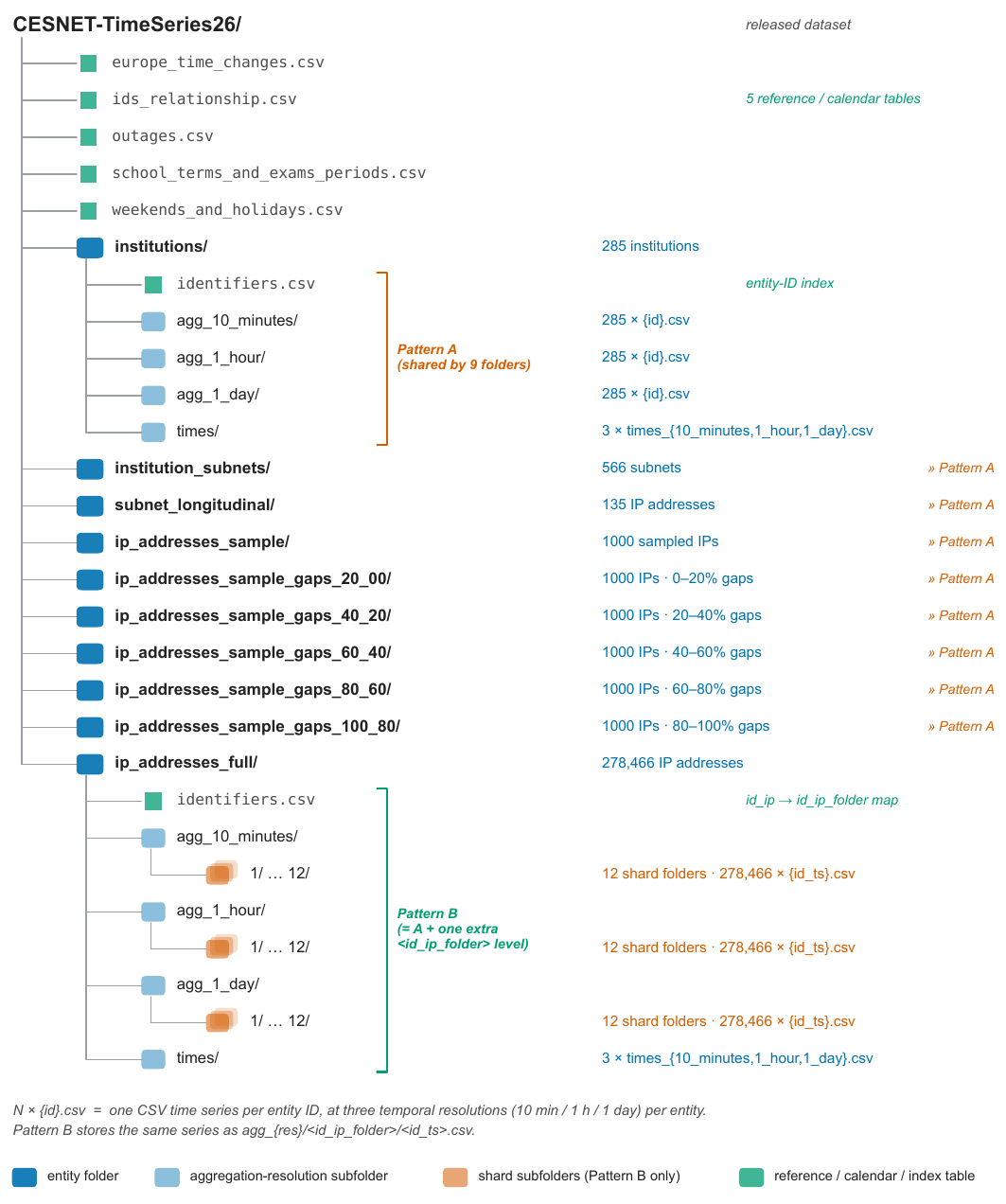}
        \caption{The file structure of the CESNET-TimeSeries26 dataset.}
        \label{fig:dataset-dirtree}
    \end{figure*}

    The dataset is delivered as compressed CSV files and is available on the Zenodo platform~\cite{CESNET_TIMESERIES26_DATASET}: \url{https://doi.org/10.5281/zenodo.21945629}. To simplify adoption, CESNET-TimeSeries26 is also integrated into the CESNET-TSZoo library~\cite{kurevs2025cesnet}: \url{https://github.com/CESNET/cesnet-tszoo}. This section describes the dataset structure and all data fields and files. The overall size of the data is $473.7\text{ GB}$ ($141\text{ GB}$ compressed).

    \subsection*{Data file hierarchy}

        The dataset structure is outlined in Figure~\ref{fig:dataset-dirtree}. Each time series type (institutional subnets, institutions, individual IP addresses, samples of IP addresses, and the longitudinal subnet) is divided by aggregation window (10 minutes, 1 hour, 1 day). Time series are stored in CSV files named by the identifier of the corresponding entity. Because identifiers are not consecutive due to filtering, each time series type has its own \texttt{identifiers.csv} file listing all included entities.

        The individual IP address (network entity) time series are further divided into subdirectories, each containing 25{,}000 files, since some file systems have difficulty handling hundreds of thousands of files in a single directory. Thus, the \texttt{identifiers.csv} also includes the subdirectory name where the corresponding series is stored.

        The dataset also contains a \texttt{times} directory for each dataset part, with translation tables that map sequence numbers in the aggregation window to absolute time. Moreover, the dataset contains a relationship table linking IP address IDs, institution IDs, and institutional subnet IDs, a table listing public holidays and weekends, a table listing school terms and exam periods, a table listing European time changes, and a table listing outages of monitoring infrastructure during the capture period.


    \subsection*{File format}

        The dataset files are organized as tables in CSV format. There are two file formats for the time series: series aggregated over 10 minutes contain the columns described in Table~\ref{tab:aggregation_metrics}, and series aggregated over one hour and one day additionally contain the columns described in Table~\ref{tab:aggregation_metrics_2}, which replace the unique-count metrics. The dataset further contains support files:

        \texttt{ids\_relationship.csv} (Table~\ref{tab:id_relationship})

        \texttt{weekends\_and\_holidays.csv} (Table~\ref{tab:weekends_and_holidays})

        \texttt{school\_terms\_and\_exams\_periods.csv} (Table~\ref{tab:school_terms})

        \texttt{europe\_time\_changes.csv} (Table~\ref{tab:europe_time_changes})

        \texttt{outages.csv} (Table~\ref{tab:outages_columns})

\section*{Technical Validation}

    This section verifies the properties of the published data in five steps. First, we check the overall properties of the collection. Second, we verify the presence of long-term multi-year behaviors. Third, we verify that repeated annual cycles are separable from anomalies. Fourth, we validate the 4.5-year longitudinal subnet. Fifth, we validate the stratified sample subsets. A closing subsection then states how the dataset relates to its 40-week predecessor. Unless stated otherwise, analyses use the 1-day aggregation, and we mask the outage days documented in Table~\ref{tab:outages} before any statistical testing.

        \begin{figure*}
            \centering
            \includegraphics[width=\linewidth]{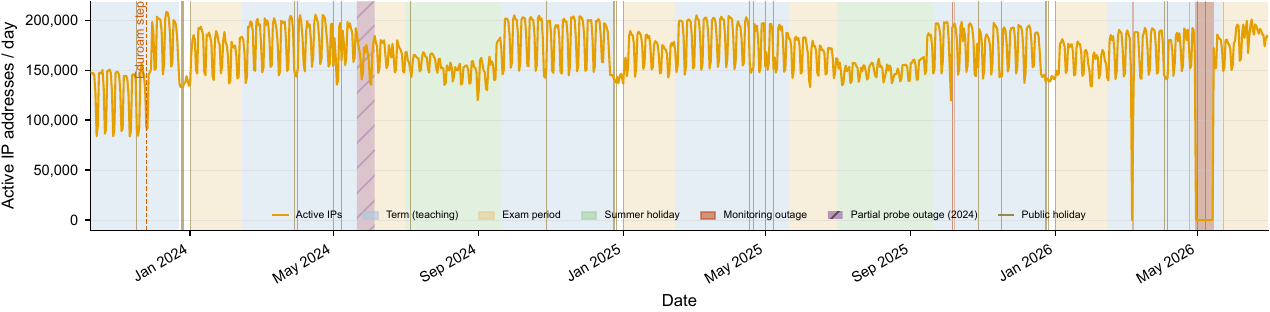}
            \caption{Daily count of active IP addresses over the 995-day collection, with known infrastructure events annotated. The November~2023 Eduroam architecture change produces a permanent level shift; the recurring Christmas and summer minima recur in every observed year.}
            \label{fig:overall_ip}

        \end{figure*}

    \subsection*{Validation of overall dataset properties}

        \begin{figure*}
            \centering
            \includegraphics[width=\linewidth]{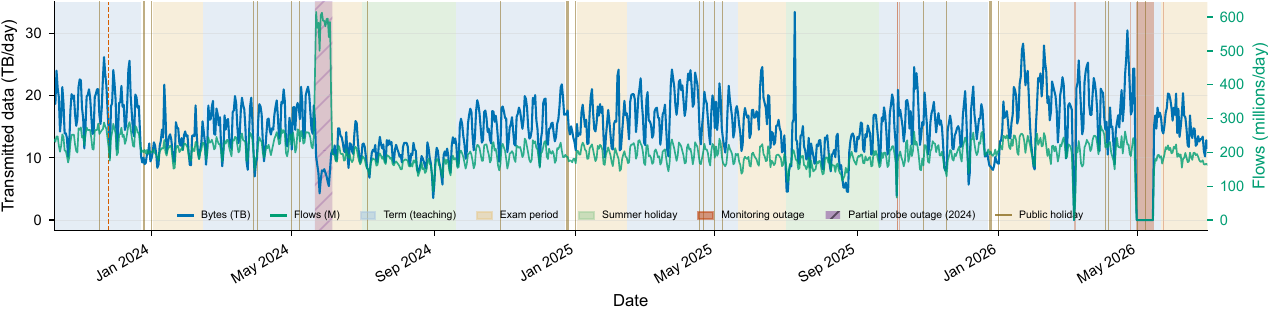}
            \caption{Daily total transmitted data over the collection. The academic calendar is directly visible, superimposed on a slow year-over-year growth trend.}
            \label{fig:overall_load}
        \end{figure*}

        \subsubsection*{Activity of IP addresses}

            Figure~\ref{fig:overall_ip} shows the daily count of active IP addresses over the full window, annotated with known events. Two observations matter for users. First, the November~2023 Eduroam public-IP architecture change was already documented in CESNET-TimeSeries24. However, now it is observable over a much longer horizon because the elevated count is sustained throughout the remaining 2.5 years and never reverts. Therefore, this ``a level shift of unknown persistence'' identified on a 40-week window is now confirmed to be a permanent architectural change rather than a transient anomaly.

            Furthermore, users of the dataset can use this labeled, expert-verified example of a genuine structural break as ground truth for evaluating change-point detectors. Second, the number of active IP addresses itself is seasonal. We can clearly see both weekly and annual seasonalities. For example, pronounced minima recur at every Christmas and every summer vacation, and the difference between teaching terms and examination periods is visible in the count alone. The practical consequence is that the population of ``present'' entities is not stationary. Therefore, per-entity analyses that implicitly assume a fixed device population (for example, cross-sectional feature distributions computed per day) will absorb this population seasonality unless they control for it.

        \subsubsection*{Evolution of transmitted data}

            The dataset is built from approximately 190.3~billion IP flows carrying 13.6~trillion packets and 13.5~petabytes of transferred data. Overall, only the IP addresses part comprises 4.5~billion 10-minute datapoint rows.  Figure~\ref{fig:overall_load} plots the daily total transmitted data (metric \texttt{n\_bytes} summed over all institutional time series). The academic rhythm dominates the profile with three autumn terms, three winter examination periods, two complete summer vacations, and three Christmas minima.

            Moreover, each annual seasonal condition recurs with the same direction and comparable depth in every observed year. We can also observe a slow secular growth of daily volume, quantified at $+0.85$~TB/day per year.

        \begin{figure*}
            \centering
            \begin{subfigure}{0.33\linewidth}
                \includegraphics[width=\linewidth]{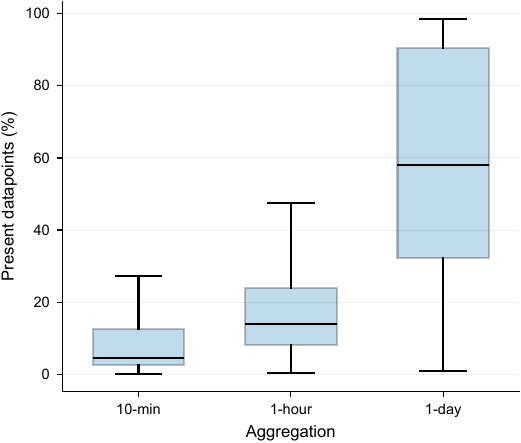}
                \caption{}
            \end{subfigure}
            \begin{subfigure}{0.33\linewidth}
                \includegraphics[width=\linewidth]{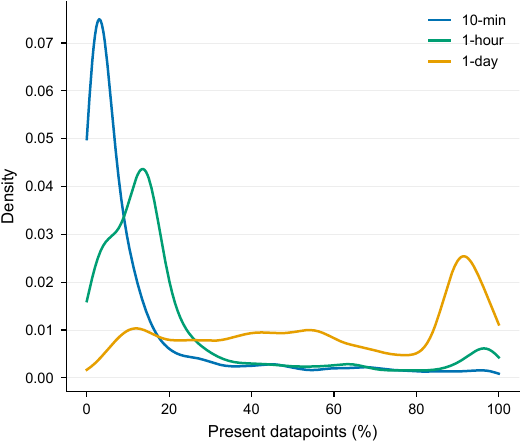}
                \caption{}
            \end{subfigure}
            \caption{Distribution of the fraction of present datapoints per series over the 995 days: (a)~boxplot by aggregation, (b)~kernel density estimate.}
            \label{fig:gaps}
        \end{figure*}

        \subsubsection*{Gaps in time series}

            As in the CESNET-TimeSeries24 dataset, most 10-minute series are sparse, as can be seen in Figure~\ref{fig:gaps}. Datapoints occupy, on average, 13.7\,\% of the 10-minute grid, with a median of only 4.6\,\%, rising to 14.0\,\% at the 1-hour aggregation and 58.1\,\% at the 1-day aggregation. The distribution is not merely skewed but bimodal, and the two modes have a physical identity.

            First mode, a sparse mode produced by end-user devices that are switched on during working hours, and a dense mode produced by always-on servers and infrastructure. Compared with the 40-week collection, the overall shape of the distribution is preserved, but both modes migrated. The peak of the sparse mode formerly located at 60--70\,\% presence now sits at 80--100\,\%, partially merging with the dense mode. The likely cause is that entities observable across 2.7 years are disproportionately permanently connected machines, while intermittently used devices accumulate absence. Two consequences follow. First, the appropriate aggregation is method-dependent: at 10 minutes, the median series is 95\,\% empty and only gap-robust methods are applicable, whereas at 1 day, the median series is majority-present and classical forecasting becomes meaningful.

            Second mode, gap severity is so heterogeneous across entities that any single random sample conflates easy and hard series. This observation motivates the stratified sample subsets validated at the end of this section.

        \subsubsection*{Entity lifetime}

            A question that only a multi-year collection can answer is how long an individual entity (IP address) remains active in network traffic. The median lifetime of the 278{,}466 IP addresses is 972 days out of the 976.6 days that carry data (the 995-day window minus the outages), with an IQR of $[924, 976]$. Thus, 58.1\,\% of entities are observable for at least 99\,\% of the window and 72.8\,\% for at least 95\,\%.

            The population decomposes into interpretable cohorts. A full-span cohort of 63.0\,\% entities is present from the first to the last day of the collection. A second cohort of 25.7\,\% entities first appears in late November~2023 during the mentioned Eduroam expansion and persists through the end. A third cohort of 7.0\,\% entities shows genuine arrivals and departures spread over time, and a short-lived tail of 2.4\,\% entities disappears within the first months of their lifetime. Two consequences follow.

            First, the dataset is well-suited to long-history per-entity modeling because more than 2.5 years of history is available for most entities. Second, lifetimes must be read as \emph{observability windows}, not as device histories, because an IP address may be reassigned to different physical devices within a single series. This can appear multiple times in a single day on heavily used Eduroam WiFi networks, which are also included in the dataset.

        \begin{figure*}
            \centering
            \begin{subfigure}{0.4\linewidth}
                \includegraphics[width=\linewidth]{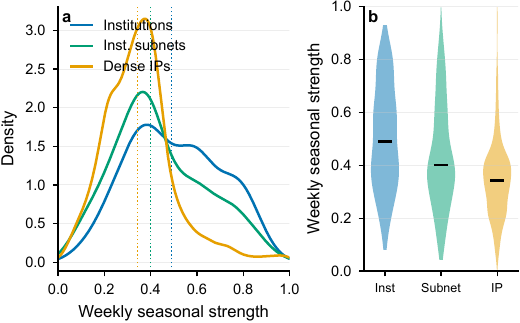}
                \caption{Weekly seasonal strength}
            \end{subfigure}
            \begin{subfigure}{0.4\linewidth}
                \includegraphics[width=\linewidth]{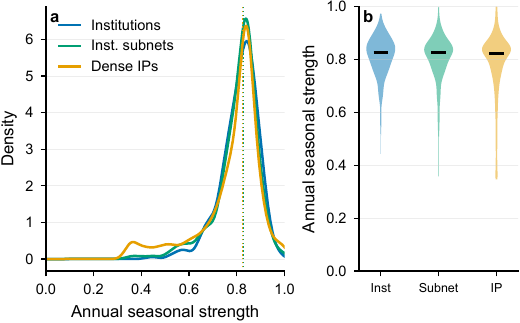}
                \caption{Annual seasonal strength}
            \end{subfigure}
            \caption{Distribution of Wang--Smith--Hyndman weekly and annual seasonal strengths across institutions, institutional subnets, and dense IP series (using metric \texttt{n\_bytes}). Annual strength (median 0.83) exceeds weekly strength (medians 0.34--0.49) for 96.8\,\% of institutions, 98.2\,\% of institutional subnets, and 99.8\,\% of dense IP series.}
            \label{fig:seasonal_atlas}
        \end{figure*}

    \subsection*{Validation of long-term structure}

        This subsection verifies the presence of long-term multi-year behaviors, which are an annual seasonal component, a calendar structure tied to the academic year, and year-over-year drift. Each is reported as a measured property with its statistical support.

        \subsubsection*{Annual seasonality is the dominant periodic component}

            We first ask which periodic component actually carries the seasonal variation. MSTL (Multiple Seasonal-Trend decomposition using Loess) splits a daily series into a trend, one seasonal component for each specified period, and a remainder. The Wang--Smith--Hyndman seasonal strength~\cite{wang2006characteristic} then measures how much of the variance that remains after removing the trend is explained by a given seasonal component. It ranges from 0 (no seasonality at that period) to 1 (the component explains all remaining variance). Using an MSTL decomposition with weekly and annual periods and the Wang--Smith--Hyndman strength measure~\cite{wang2006characteristic} on daily transmitted data (metric \texttt{n\_bytes}), we compute both strengths for every entity with sufficient coverage. The distribution of Wang--Smith--Hyndman strength is shown in Figure~\ref{fig:seasonal_atlas}.

            The median annual strength is 0.83 at every entity level, whereas the median weekly strength ranges only from 0.34 (dense IP series) to 0.49 (institutions). Moreover, the annual component is stronger than the weekly one for 96.8\,\% of 280 institutions, 98.2\,\% of 544 institutional subnets, and 99.8\,\% of 5{,}000 randomly selected dense IP series (IP address with maximum 40\,\% of gaps). In other words, the strongest regular signal in academic-network traffic is not the familiar weekly rhythm on which sub-year benchmarks concentrate, but the annual one.

            Week-to-week seasonal changes are caused by timetables changing between the halves of a term, by attendance declining as a term progresses, and by examination weeks and short breaks interrupting the weekly routine. By contrast, the annual cycle is fixed by the academic calendar, which places terms, examination periods, and vacations at nearly the same position every year. Therefore, the weekly component is less regular than the annual one, and the strength measure reflects that difference. The observed dominance of the annual component is thus consistent with how the network is used.

        \begin{figure*}
              \centering
              \begin{subfigure}{0.4\linewidth}
                  \includegraphics[width=\linewidth]{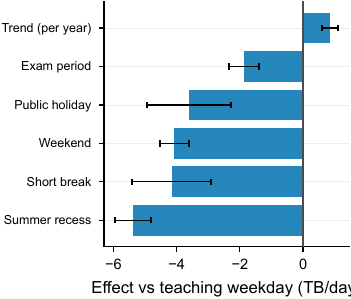}
                  \caption{Effect of each calendar condition on daily transmitted volume relative to a teaching weekday, with 95\,\% confidence intervals. The conditions order by campus occupancy rather than by whether teaching is formally in session.}
                  \label{fig:calendar-a}
              \end{subfigure}
              ~
              \begin{subfigure}{0.4\linewidth}
                  \includegraphics[width=\linewidth]{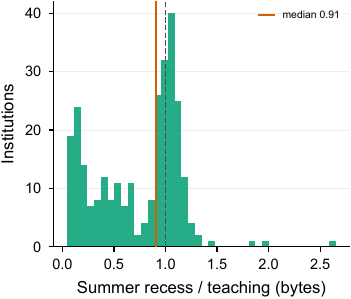}
                  \caption{Distribution of the summer-recess-to-teaching volume ratio computed separately for each institution. This analysis revealing two populations (universities and office-like networks) that the network aggregate blends into a single curve.}
                  \label{fig:calendar-c}
              \end{subfigure}
              \caption{Calendar structure of the network load, measured directly by conditioning on the academic calendar.}
              \label{fig:calendar}
          \end{figure*}

        \subsubsection*{Load tracks campus occupancy, not the teaching calendar}

            To verify that the calendar events correspond to the behavior of the series, we regressed daily transmitted volume on the academic regime (teaching term, examination period, short break, summer recess), weekends, and public holidays. Figure~\ref{fig:calendar-a} shows the shift in daily volume relative to a teaching weekday, with a fitted level of $16.74$~TB/day. The trend term is a slope rather than a level shift and is reported separately: overall daily volume grows by $0.85$~TB/day per year. The conditions fall into three tiers rather than a strict ordering. Summer recess removes the most traffic. Short breaks, weekends, and public holidays form a middle group whose coefficients are mutually indistinguishable. Examination periods remove the least, and are separated from every other condition.

            That grouping carries the substantive finding that the network does not respond to whether teaching is formally in session but to how many people are physically on campus. Examination periods keep buildings, laboratories, and dormitories occupied and retain roughly two-thirds of the volume that a summer recess loses, whereas a public holiday is not measurably different from an ordinary weekend. Two properties of the data follow. First, weekends and public holidays are statistically indistinguishable in their effect on volume, so the data do not support treating them as separate regimes. Second, the calendar conditions and the trend together explain $R^2 = 0.437$ of the daily variance. The remaining $56\,\%$ is the variation that the calendar does not account for, and it is the part of the signal on which anomaly detection operates.

        \subsubsection*{The aggregate academic calendar effect is a property of a few large institutions, not of the network}

            Computing the summer-recess-to-teaching volume ratio separately for each of the $280$ institutions with sufficient data yields a median of $0.909$ and an IQR of $[0.366, 1.052]$. The distribution is distinctly bimodal, as shown in Figure~\ref{fig:calendar-c}. The  $32.9\,\%$ institutions fall below $0.5$ of their teaching-period volume, $35.0\,\%$ sit at or above $1.0$, and only $15.7\,\%$ occupy the intermediate range $[0.5, 0.9)$ that the aggregate coefficient would suggest is typical. To test whether recess changes an institution's volume at all, we applied a two-sided Mann--Whitney $U$ test to the daily volumes of recess days versus teaching days of each institution. The test asks whether the two sets of days come from the same distribution and makes no assumption about its shape. It finds the teaching-versus-recess contrast significant for $68.9\,\%$ institutions.

            However, significance is not synonymous with a summer decline: $51.8\,\%$ of those institutions decrease, and $17.1\,\%$ significantly \emph{increase} their traffic during recess. The largest confirmed increases are expected, given the institution's type. The two strongest are in the health sector, whose operations do not follow the academic calendar. The next is a national Internet infrastructure operator whose traffic is not driven by on-campus users. The remaining large increases come from the largest server-heavy address spaces in the dataset, where maintenance, migrations, and batch computation are scheduled when users are absent.

            Transmitted volume is extremely concentrated because six institutions account for half of all traffic. Moreover, every one of the six largest falls in the seasonal mode, with recess ratios between $0.41$ and $0.90$. The two modes correspond to institution type rather than institution size, separating occupancy-driven entities such as teaching faculties and dormitories from office networks and machine-driven entities such as research clusters and hosting infrastructure, whose load is indifferent to the academic calendar. This is the clearest argument in our validation for publishing per-entity series rather than aggregates alone. An aggregate seasonal component averages two populations into a single curve that describes neither, and evaluating a seasonal model only on the network total would report a level of calendar predictability that does not transfer to the entities users actually deploy models on.

        \subsubsection*{The annual component repeats across cycles}

            Dominance of the annual component does not by itself show that it repeats. Therefore, we tested repetition on two deliberately contrasting institutions: institution~74, a university-like network, and institution~101, an office-like network. Correlating the daily profile of the first annual cycle with that of the second while shifting the phase, the university's profile peaks sharply at a $+2$-day shift with Pearson $r = 0.825$, whereas the office's correlation has no seasonal peak. Fitting the calendar effects independently in each cycle, on the decimal logarithm of daily transmitted bytes, gives a vacation effect of $-1.01$, $-0.88$, and $-0.80$ log units for the university and a public-holiday effect of $-1.52$, $-1.53$, and $-1.36$ log units for the office. Therefore, both institutions carry repeatable annual structure, but of different kinds, and the repetition is a measurable property rather than a visual impression.

        \begin{figure*}
            \centering
            \includegraphics[width=\linewidth]{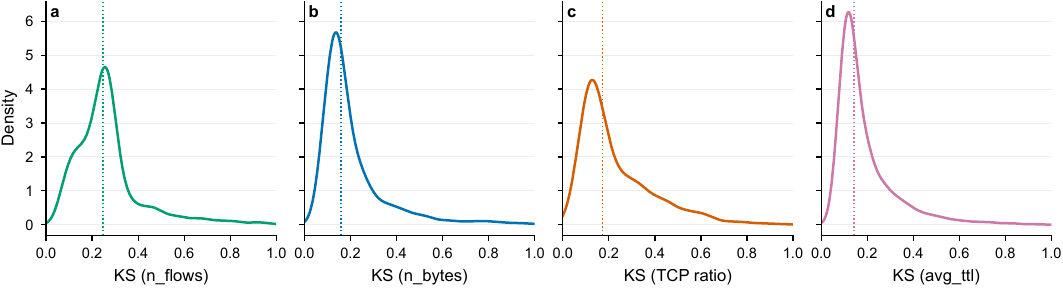}
            \caption{Distribution across entities of the year-1-versus-year-2 KS statistic for four metrics. Significant distributional drift is the norm rather than the exception (63.8--82.6\,\% of entities depending on the metric).}
            \label{fig:drift}
        \end{figure*}

        \subsubsection*{Year-over-year behavioral drift}

            Seasonality describes what repeats. Drift describes what does not. To verify that the dataset contains measurable drift, we compared the per-entity distribution of each metric in the first collection year with that in the second, using the two-sample Kolmogorov--Smirnov (KS) test over the 3{,}541 entities sufficiently observed in both years. The KS test asks whether two samples come from the same distribution; its statistic is the largest difference between the two empirical distribution functions. Significant drift ($p<0.05$) is widespread across the dataset and affects 82.6\,\% of entities for \texttt{n\_flows} (median KS statistic 0.25), 73.8\,\% for \texttt{tcp\_udp\_ratio\_bytes}, 71.5\,\% for \texttt{n\_bytes}, and 63.8\,\% for \texttt{avg\_ttl}, as shown in Figure~\ref{fig:drift}. The drift has a direction. Within this cohort, the mean daily flow count fell by roughly a quarter and the mean daily byte volume by 13\,\% from year one to year two. For most entities, next year's data is therefore measurably different from this year's.

    \subsection*{Validation of separation of annual events and anomalies}

\begin{figure*}
            \centering
            \includegraphics[width=\linewidth]{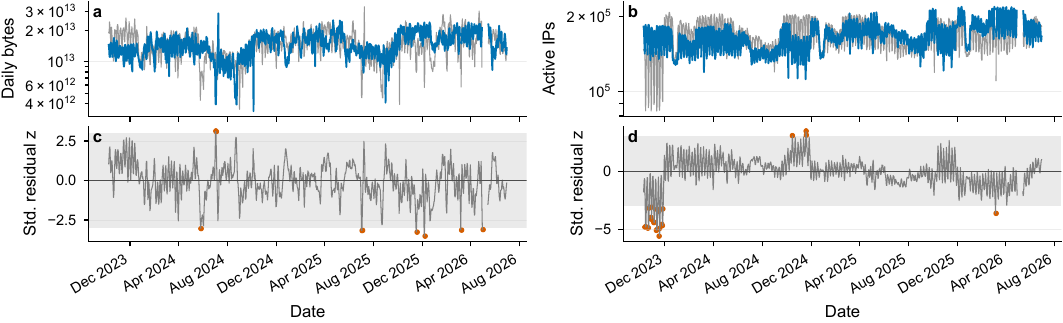}
            \caption{Leave-one-cycle-out annual baseline. Recurring calendar effects (Christmas, summer) are absorbed (median $|z|\approx0.5$--$1.2$), while non-recurring events remain detectable (2024 probe outage $|z|\approx3.0$ in byte volume. November~2023 Eduroam step $|z|\approx5.6$ in the active-IP count).}
            \label{fig:calendar_anomaly}
        \end{figure*}

        Calendar structure poses a specific problem for anomaly detection. A Christmas traffic collapse is a large deviation from the preceding weeks. A detector trained on less than a year of data must either raise an alarm on it or be told manually to ignore it. Repeated annual cycles resolve this problem in a principled way. We formed a leave-one-cycle-out annual baseline, in which the expectation for each cycle is built only from the other cycles, and scored standardized robust residuals against it, as shown in Figure~\ref{fig:calendar_anomaly}.

        The separation works in both directions. The baseline absorbs the recurring calendar effects. The Christmas and summer periods, which would dominate the alarm list of any short-window detector, collapse to median residuals of $|z|\approx0.5$--$1.2$. Non-recurring events remain prominent. The 2024 partial probe outage reaches $|z|\approx3.0$ and the November~2023 Eduroam architecture step $|z|\approx5.6$. The two events surface in different metrics. The outage is most visible in byte volume, whereas the Eduroam step is most pronounced in active-IP count. Therefore, the dataset supports the evaluation of seasonally aware anomaly detectors with a built-in ground-truth contrast: two labeled, non-recurring events that a detector must flag, and a set of recurring calendar effects that it must not flag.

    \subsection*{Validation of the longitudinal subnet}

        \begin{figure*}[t]
            \centering
            \includegraphics[width=\linewidth]{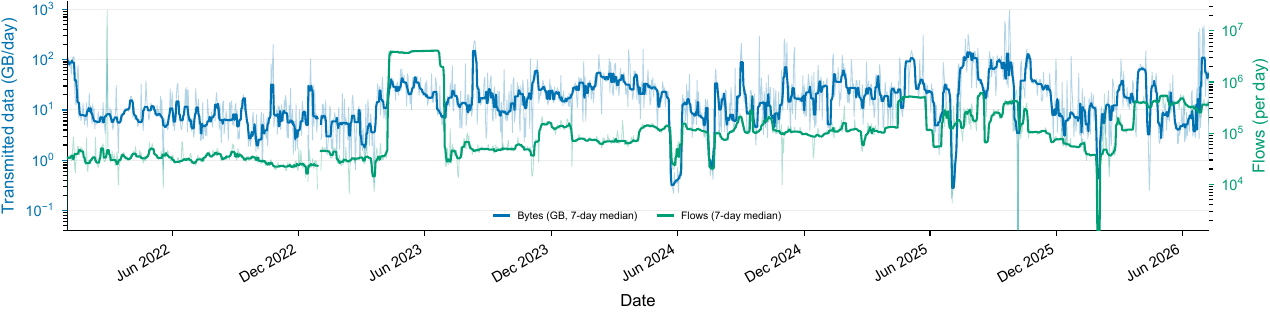}
            \caption{Daily total transmitted data and flow count of the longitudinal subnet over its 4.5-year span (7-day medians). The elevated flow count between April and July 2023 is driven by a single short-lived address (IP identifier 1590672), which appears in the dataset only from 2023-04-09 to 2023-07-17 and accounts for 99.5\,\% of the excess flows during that period.}
            \label{fig:overall_load_subnet}
        \end{figure*}

        The 4.5-year longitudinal subnet extends the analyses above beyond the reach of the main collection, as shown in Figure~\ref{fig:overall_load_subnet}. One feature of the figure needs a note before the findings. Between April and July 2023, the summed flow count increases by about one order of magnitude, then returns to its previous level. This is not a measurement artifact. A single address, IP identifier 1590672, accounts for 99.5\,\% of the excess flows in that period, and the five most active addresses together account for 99.8\,\%. That address is present in the dataset only from 2023-04-09 to 2023-07-17, so it appears and disappears together with the excess. Therefore, the episode is a genuine short-lived entity, visible in the per-address series, and it is one example of why the subnet aggregate must be read together with its member series. We keep it in the data and in the figure as a labeled event.

        Three findings follow from the subnet's length. First, its traffic grows strongly. Summed over the 135 series, byte volume rises by roughly $+277\,\%$ and flow counts by $+775\,\%$ over the 4.5 years. However, it is caused by two IP addresses from all of them. Without the series with ID 79658, it is $+88\,\%$, and also without the series with ID 75864, it is $+26\,\%$ during the whole period.
        
        Second, the two volumetric metrics behave differently across cycles. The summed byte profile does not reproduce annually (mean cross-cycle Pearson $r\approx-0.1$) because the byte volume of a small set of entities is heavy-tailed, and a few large transfers can dominate any given week. Flow counts track the number of connections rather than their size, and they exhibit a weak, recurring academic rhythm. 
        
        Third, the four complete cycles allow an evaluation that, to our knowledge, no other public network dataset supports: training on three full cycles and forecasting the entire held-out fourth year. On flows, a weekly-plus-annual model beats a weekly-only model over the full-year horizon (SMAPE 53.2 versus 60.7, $R^2$ $+0.15$ versus $-0.05$). On bytes, the annual component brings nothing (SMAPE 81.6 versus 81.5, $R^2\leq0$). At this small aggregation scale, annual structure is therefore usable for count metrics but not for volume metrics.

        \begin{figure*}[t]
            \centering
            \includegraphics[width=0.9\linewidth]{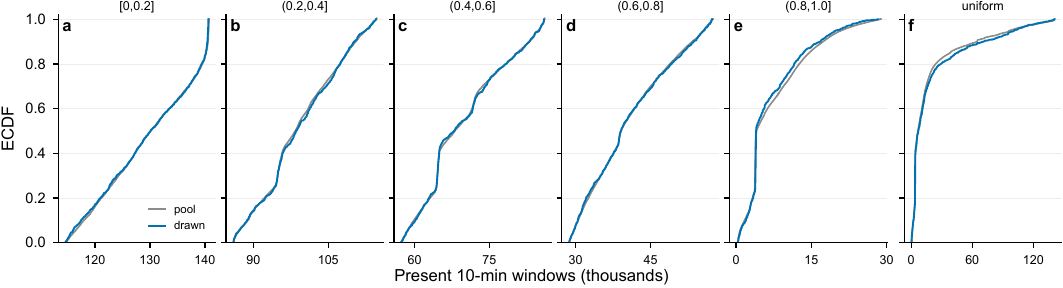}
            \caption{Per-band comparison of the \texttt{n\_present} distribution of the drawn 1{,}000 entities against the full band candidate pool. Distributions coincide across all bands (KS $p=0.12$--0.93), confirming that each subset is a representative draw from its band.}
            \label{fig:sample_integrity}
        \end{figure*}

    \subsection*{Validation of the stratified sample subsets}

        The full IP address part contains more than 278 thousand series. Evaluating a method on all of them is impractical, and a single random sample reports only one average difficulty. Therefore, we publish six fixed-size ($N=1000$) sample subsets: a uniform-random baseline that reproduces the original CESNET-TimeSeries24 sampling, and five subsets stratified by percentage of gaps into equidistant windows. We validate the subsets for stratification integrity and representativeness, so that authors can select a subset with confidence in its contents.

        Every subset is exactly sized (1{,}000 drawn from pools that all exceed 1{,}000) with zero out-of-band entities. Deciding membership in integer \texttt{n\_present} space removes float ambiguity entirely. Within each band, the drawn 1{,}000 are statistically indistinguishable from the full band pool in terms of the number of present datapoints, as shown in Figure~\ref{fig:sample_integrity}. The uniform baseline is indistinguishable from the entire 278{,}466-entity population. Therefore, the baseline can stand in for the dataset as a whole. The five bands are stratified by design and must \emph{not} be concatenated into a pseudo-representative pool, because doing so would weight sparse series at one two-hundredth of their true prevalence. The candidate pool sizes themselves are a finding about the population. The very-sparse band draws from 228{,}387 entities (82.0\,\% of the dataset), whereas the dense $[0.00,0.20]$ band draws from only 7{,}852 (2.8\,\%), and the three densest bands each hold fewer than 13{,}000 candidates. Dense, always-on series, the kind on which forecasting papers prefer to demonstrate, are a small minority of real ISP-scale data.

\subsection*{Relationship to CESNET-TimeSeries24}

    CESNET-TimeSeries26 is the successor of CESNET-TimeSeries24~\cite{koumar2025cesnet}, not an extension of it. This section explains how the two datasets relate and what the changes are over the multi-year span. We use CESNET-TimeSeries24 as the controlled instance of a sub-year collection because it comes from the same network, monitoring pipeline, and metrics, and differs only in length. The conclusions apply to sub-year network traces in general.

    \begin{table*}[t]
            \centering
            \caption{\label{tab:overlap}Relationship of the CESNET-TimeSeries26 IP-address entity set to CESNET-TimeSeries24, under the recomputed relative filtering threshold ($>99.75\%$ empty, i.e., $\leq358$ non-empty 10-minute datapoints).}
            \rowcolors{3}{gray!10}{white}
            \begin{tabular}{l|r|r|r}
                  \textbf{Entity set} & \textbf{Institutions} & \textbf{Institution subnets} & \textbf{IP addresses} \\
                  \hline
                  CESNET-TimeSeries24 entities & 283 & 548 & 275{,}124 \\
                  CESNET-TimeSeries26 entities  & 285 & 566 & 278{,}466 \\
                  Shared (qualify in both) & 283 & 548 & 253{,}543 \\
                  Added (qualify now, absent before) & 2 & 18 & 24{,}923 \\
                  Dropped (in TS24, no longer qualify) & 0 & 0 & 21{,}581 \\
                  \hline
            \end{tabular}
        \end{table*}

    \paragraph{Continuity} The capture pipeline, metrics, aggregation, and anonymization are unchanged, and the anonymized identifiers denote the same entities in both datasets. Table~\ref{tab:overlap} shows overlaps between entities in both datasets. Of the 275{,}124 IP-address series of CESNET-TimeSeries24, 253{,}543 remain in the new dataset under the recomputed relative completeness threshold, 21{,}581 no longer qualify, and 24{,}923 series that became active during the extension are new. Furthermore, all 283 institutions and 548 institutional subnets are retained, with 2 and 18 added. Therefore, results, labels, and identified events published for the 40-week dataset remain valid starting points for this one. Moreover, every series in CESNET-TimeSeries26 whose ID also appears in CESNET-TimeSeries24 is a strict extension of it, i.e., its first 40 weeks are identical to the corresponding series in CESNET-TimeSeries24..   

    \begin{figure*}[t]
        \centering
        \includegraphics[width=\linewidth]{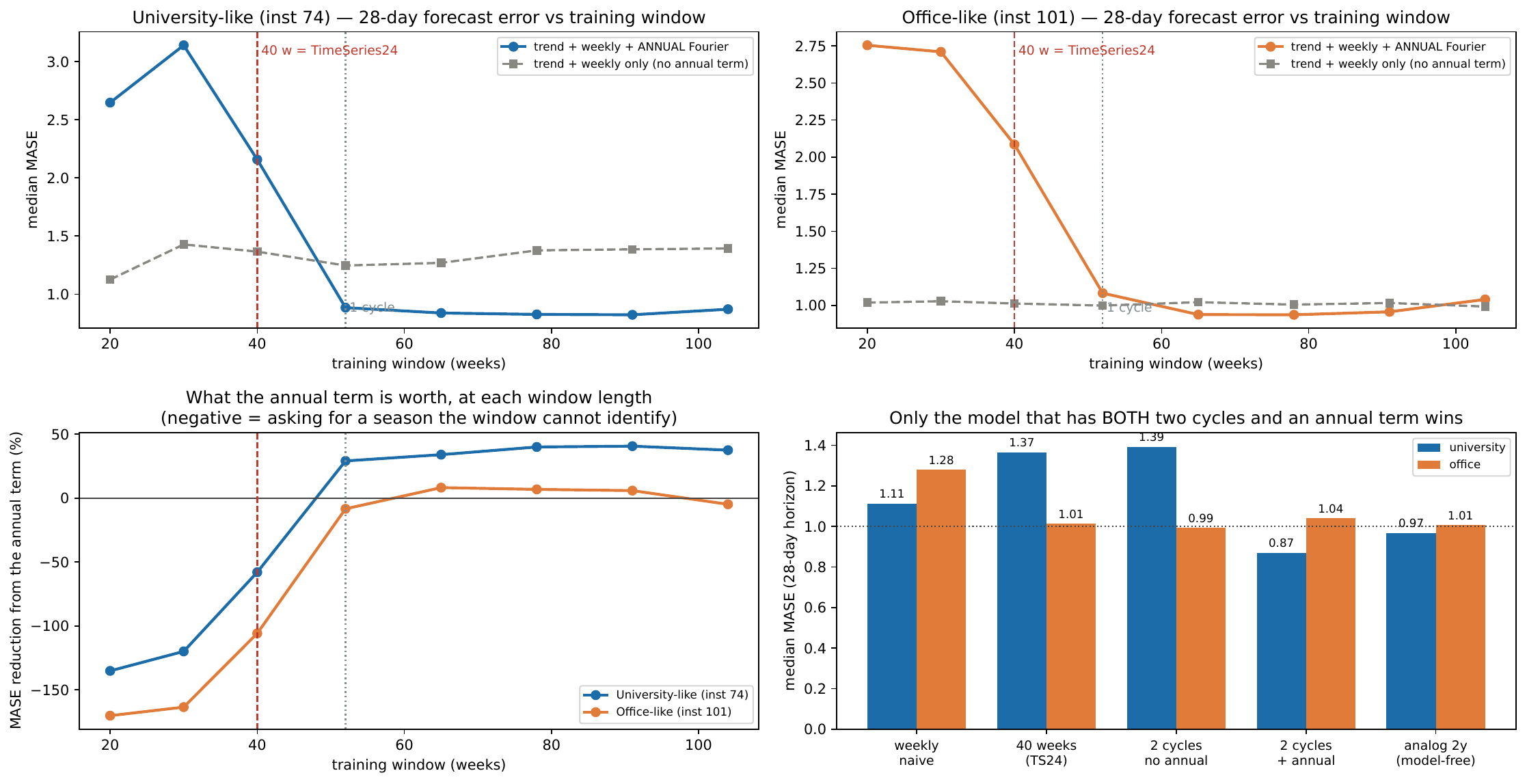}
        \caption{Exploitability of the annual component in a rolling-origin backtest with a 28-day horizon. Top: median MASE as a function of training-window length, with and without the annual term. The annual model starts to win only once the window covers one full cycle. Bottom left: the MASE reduction attributable to the annual term for each window length. Below one cycle, the contribution is negative. Bottom right: only the model that includes both historical cycles and an annual term beats all baselines for the university. The office is served equally well by any weekly model.}
        \label{fig:annual_case_exploit}
    \end{figure*}

     \paragraph{The dominant periodic component is unobservable below one cycle} The annual component carries more of the periodic variation than the weekly one for 96.8\,\% of institutions, 98.2\,\% of institutional subnets, and 99.8\,\% of dense IP series, as shown in Figure~\ref{fig:seasonal_atlas}. A record shorter than one year cannot contain this component. A test on the university-like institution~74 places the break-even point for an annual term at exactly one complete cycle, as shown in Figure~\ref{fig:annual_case_exploit}. Below that length, the annual term is harmful because a sub-year window observes only a fragment of the annual curve, thus, the seasonal and trend components cannot be separated from each other. 
     
     A harmonic model trained on a 40-week window, the span of CESNET-TimeSeries24, reaches a median MASE of $1.37$. The same model class trained on two full cycles with an annual term reaches $0.87$. Two cycles of history \emph{without} the annual term yield $1.39$, similar as for the 40-week window. Therefore, a longer history is not valuable in itself. It becomes valuable when the model can express the annual component. Across all 255 institutions with sufficient coverage, the gain from the annual term grows with the depth of the institution's summer drop (Spearman $\rho=-0.571$, $p=2\cdot10^{-23}$). This is expected observation because only a model with an annual term can forecast correctly the behavior change from the teaching period to the summer break. The mayority of institutions are universities and schools for which is change of behavior at the start of summer holidays significant behavior.

    \paragraph{Growth and seasonality separate only over multiple years} The fitted trend of $+0.85$~TB/day per year ($+5.1\,\%$) is small against a summer-recess drop of $5.38$~TB/day. Over the 40-week window of CESNET-TimeSeries24, the trend accumulates to roughly $0.65$~TB/day, which is less than the uncertainty in the seasonal coefficients. Over 2.7 years, it accumulates to about $2.3$~TB/day and separates cleanly. A sub-year record cannot distinguish secular growth from the rising limb of a seasonal cycle, so any capacity-planning or long-horizon forecasting study built on it extrapolates a quantity it cannot identify.

    \paragraph{Calendar effects cannot be separated from anomalies without a second cycle} The leave-one-cycle-out baseline of Figure~\ref{fig:calendar_anomaly} needs at least two complete cycles, because the expectation for each cycle must come from other cycles. On a sub-year record, a detector has no annual expectation to violate; it must either alarm on every Christmas and every summer recess or be told manually to ignore them. This class of evaluation, seasonally aware anomaly detection with a built-in contrast between recurring and non-recurring events, is therefore unavailable on any sub-year dataset.

\begin{figure*}
        \centering
        \includegraphics[width=0.6\linewidth]{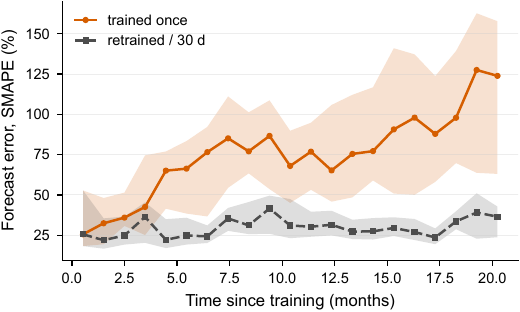}
        \caption{Forecast error versus time since training. Without retraining, median SMAPE grows by $\sim$11.3 points per quarter (25.5\,\%~$\rightarrow$~123.9\,\% over 20 months); retraining every 30~days holds error near 29.5\,\%, implying a practical retraining interval of roughly two to three months.}
        \label{fig:decay}
    \end{figure*}

    \paragraph{Model staleness is measurable only over multiple years} The drift verified above has an operational consequence: a deployed model goes stale. To measure how fast, we trained a fixed seasonal model once on the first annual cycle and evaluated it, without retraining, on consecutive 30-day blocks over the following 20 months, on 40 dense series (20 institutions and 20 IP addresses). We compared it against the same model class retrained every 30 days on a trailing one-cycle window. The stale model's median SMAPE grows from 25.5\,\% to 123.9\,\%, approximately 11.3~SMAPE points per quarter, whereas the retrained model stays near 29.5\,\% (Figure~\ref{fig:decay}). After about 2.3~months the stale error exceeds $1.5\times$ the fresh baseline, which implies a practical retraining interval of roughly two to three months for this model class. Measuring this decay requires a complete training cycle followed by a long evaluation period. A sub-year record cannot even complete the training cycle, so the staleness of a deployed model is, to our knowledge, measurable on public network data for the first time.

\section*{Usage Notes}

    This section describes recommendations for evaluating network traffic forecasting (and forecasting-based anomaly detection) on this dataset to ensure that results from different authors remain comparable. The checklist is shown in Table~\ref{tab:recommendation-checklist}.

    \begin{table*}
        \centering
        \caption{Checklist for addressing recommended evaluation procedures.}
        \label{tab:recommendation-checklist}
        \begin{tabular}{c|p{15cm}}
            \textbf{No.} & \textbf{Recommendation} \\
            \hline
            \hypertarget{row1}{(1)} & Specify which dataset part(s) are used in the analysis. \\
            \rowcolor[HTML]{EFEFEF} \hypertarget{row2}{(2)} & Specify the aggregation interval(s) used. \\
            \hypertarget{row3}{(3)} & Indicate whether the approach is multivariate or univariate. \\
            \rowcolor[HTML]{EFEFEF} \hypertarget{row4}{(4)} & Clearly state if not all metrics are used. \\
            \hypertarget{row5}{(5)} & Document all preprocessing steps, including filtering, normalization, and gap handling. \\
            \rowcolor[HTML]{EFEFEF} \hypertarget{row6}{(6)} & Ensure the training phase starts from the beginning of the relevant part's time frame (main dataset: 2023-10-09; longitudinal subnet: its own 2022 start). \\
            \hypertarget{row7}{(7)} & Specify the duration of the training window. \\
            \rowcolor[HTML]{EFEFEF} \hypertarget{row8}{(8)} & Define and describe the validation window if employed. \\
            \hypertarget{row9}{(9)} & Clearly describe the retraining process if the model is retrained during evaluation. \\
            \rowcolor[HTML]{EFEFEF} \hypertarget{row10}{(10)} & Specify the forecasting horizon. \\
            \hypertarget{row11}{(11)} & Clearly specify the evaluation metrics used. \\
            \rowcolor[HTML]{EFEFEF} \hypertarget{row12}{(12)} & Provide an overall comparison across series using statistical distributions and aggregate statistics. \\
            \hypertarget{row13}{(13)} & Assess and document computational requirements and deployability. \\
            \rowcolor[HTML]{EFEFEF} \hypertarget{row14}{(14)} & Make source code of experiments and models publicly available. \\
            \hypertarget{row15}{(15)} & Report results separately per evaluation year (or clearly delimited sub-periods) in addition to overall results, so the effect of behavioral drift is visible. \\
            \rowcolor[HTML]{EFEFEF} \hypertarget{row16}{(16)} & When claiming annual-seasonality modeling, ensure the training window covers at least one complete annual cycle and state this explicitly. \\
            \hypertarget{row17}{(17)} & Prefer the canonical split for comparability: train 2023-10-09 to 2025-10-08 (two full years), test the remainder. \\
            \hline
        \end{tabular}
    \end{table*}

    \paragraph{Dataset selection} The dataset is divided into several parts, each of which is independently usable: Full IP addresses, Sample IP addresses, Sample IP addresses with exact gap ratio, Institutions, Institutional subnets, and Longitudinal subnet. Authors must state which part(s) \hyperlink{row1}{(1)} and which aggregation interval(s) \hyperlink{row2}{(2)} they use, reporting results separately for each. The approach must indicate whether it is multivariate or univariate \hyperlink{row3}{(3)}, and any unused metrics must be stated \hyperlink{row4}{(4)}. All preprocessing must be described \hyperlink{row5}{(5)}. Given that the Full IP address part contains more than 278 thousand series, evaluating all of them is challenging. Therefore, we also publish a stratified sample subset and a stratified sample subsets with predefined gap ratios.

    \paragraph{Longitudinal subnet caveats} Three caveats apply to the longitudinal subnet part. It is TCP-only throughout, so its absolute volumetric levels must not be compared directly with the all-protocol parts, and it must not be used in analyses that depend on UDP traffic. Its transport-ratio metrics are constant and should be excluded from the model. Recommendation~\hyperlink{row6}{(6)} applies with the part's own 2022 start date rather than 2023-10-09.
    
    \paragraph{Training and forecasting correctness} The training phase must start at the beginning of the relevant part's time frame \hyperlink{row6}{(6)}, with the training-window duration stated \hyperlink{row7}{(7)}; a validation window, if used, must be defined \hyperlink{row8}{(8)}; retraining must be described \hyperlink{row9}{(9)}; and the forecasting horizon must be specified \hyperlink{row10}{(10)}.

    \paragraph{Evaluation metrics} The chosen metrics must be specified \hyperlink{row11}{(11)}. We recommend the following ($n$ observations, $y_i$ actual, $\hat{y}_i$ predicted):
    \begin{itemize}
        \item \textbf{Root Mean Squared Error (RMSE)}, sensitive to large errors~\cite{chicco2021coefficient}:
        \begin{equation}
        \text{RMSE} = \sqrt{\frac{1}{n} \sum_{i=1}^n (y_i - \hat{y}_i)^2} \label{rmse}
        \end{equation}
        For combining multiple RMSEs we recommend the weighted RMSE:
        \begin{equation}
        \text{Weighted RMSE} = \sqrt{\frac{\sum_{i=1}^{n} \sigma_i^2 \cdot \text{RMSE}_i^2}{\sum_{i=1}^{n} \sigma_i^2}}
         \label{rmse-weighted}
        \end{equation}
        where $\sigma_i^2$ is the variance of the true values in the $i$-th series.
        \item \textbf{Symmetric Mean Absolute Percentage Error (SMAPE)}~\cite{chicco2021coefficient}, with $\epsilon$ a small constant:
        \begin{equation}
        \text{SMAPE} = \frac{100\%}{n} \sum_{i=1}^n \frac{|y_i - \hat{y}_i|}{(|y_i| + |\hat{y}_i| + \epsilon) / 2}  \label{smape}
        \end{equation}
        with weighted variant
        \begin{equation}
        \text{Weighted SMAPE} = \frac{\sum_{i=1}^{n} \sigma_i^2 \cdot \text{SMAPE}_i}{\sum_{i=1}^{n} \sigma_i^2}
         \label{smape-weighted}
        \end{equation}
        \item \textbf{Coefficient of Determination ($R^2$)}~\cite{wright1921correlation,chicco2021coefficient}:
        \begin{equation}
            R^2 = 1 - \frac{\sum_{i=1}^n (y_i - \hat{y}_i)^2}{\sum_{i=1}^n (y_i - \mu_y)^2}   \label{r2score}
        \end{equation}
        with weighted variant
        \begin{equation}
            \text{Weighted } R^2 = \frac{\sum_{i=1}^{n} \sigma_i^2 \cdot R_i^2}{\sum_{i=1}^{n} \sigma_i^2}
            \label{r2score-weighted}
        \end{equation}
    \end{itemize}

    \paragraph{Multiple time series evaluation} All metrics are computed per series; an overall comparison across series must additionally be made \hyperlink{row12}{(12)} using distributions per metric (mean, standard deviation, the weighted variants, and histogram/KDE plots).

    \paragraph{Computational requirements and deployability} Evaluation should assess not only accuracy but also training time, inference speed, and resource consumption \hyperlink{row13}{(13)}; excessive requirements may preclude real-time or resource-limited deployment. The source code of experiments and models should be made publicly available \hyperlink{row14}{(14)}.

    \paragraph{Per-year reporting and annual-cycle claims} Because the dataset spans multiple years and the behavior of most entities drifts from year to year, an overall metric can mask a model that degrades over time. Results should therefore be reported per evaluation year or per clearly delimited sub-period \hyperlink{row15}{(15)}. Any approach claiming to model annual seasonality must use a training window that covers at least one complete annual cycle and state this explicitly \hyperlink{row16}{(16)}. Below that length, the annual term degrades the forecast. For cross-study comparability, we recommend the canonical split \hyperlink{row17}{(17)}: train on the first two full years (2023-10-09 to 2025-10-08) and test on the remainder.

\section*{Code Availability}

\begin{table*}
    \centering
    \footnotesize
    \caption{Software used for creating the dataset.}
    \begin{tabular}{l|c|l}
         \textbf{Name} & \textbf{Version} & \textbf{Link}  \\
         \hline
         Ipfixprobe & 4.11.0 & \url{https://github.com/CESNET/ipfixprobe} \\
         \rowcolor[HTML]{EFEFEF} IPFIXcol2 & 2.2.1 & \url{https://github.com/CESNET/ipfixcol2} \\
         NEMEA Framework & 0.14.0 & \url{https://github.com/CESNET/Nemea-Framework} \\
         \rowcolor[HTML]{EFEFEF} NEMEA modules & 2.20.0 & \url{https://github.com/CESNET/Nemea-Modules} \\
         NEMEA Supervisor & 1.8.2 & \url{https://github.com/CESNET/Nemea-Supervisor} \\
         \rowcolor[HTML]{EFEFEF} TimeScaleDB-14 & 2.15.0 & \url{https://www.timescale.com/} \\
         Python  & 3.9.0 & \url{https://www.python.org/downloads/release/python-390/} \\
         \rowcolor[HTML]{EFEFEF} Create Datapoints module & -- & \url{https://github.com/koumajos/CESNET-TimeSeries24-CD} \\
         \hline
    \end{tabular}

    \label{tab:softwareversion}
\end{table*}

The dataset has been produced using open-source software: the flow exporter Ipfixprobe, the flow collector IPFIXcol2, the NEMEA processing system and modules (all on GitHub), and the TimeScaleDB database. We also provide the Create Datapoint module and deployment scripts for the NEMEA Supervisor and for building the database. Software versions and repository links are summarized in Table~\ref{tab:softwareversion}. 

\paragraph{Integration into CESNET-TSZoo} To simplify adoption, CESNET-TimeSeries26 is also integrated into the CESNET-TSZoo library~\cite{kurevs2025cesnet}: \url{https://github.com/CESNET/cesnet-tszoo}. This library already provides access to CESNET-TimeSeries24 dataset and others network traffic time series datasets. Therefore, researchers can also download the dataset using the library, select the aggregation level and the source type (IP addresses, institutions, or institutional subnets), and create train, validation, and test splits through the same interface they already use for CESNET-TimeSeries24 dataset. Because the library handles data loading, filtering, and preprocessing in a consistent way, experiments can move between the two datasets with minimal code changes.


\section*{Acknowledgements}

    This research was supported by the Ministry of Education, Youth and Sports of the Czech Republic in the project ``e-Infrastructure CZ'' (LM2023054). Computational resources were provided by the e-INFRA CZ project (ID:90254), supported by the Ministry of Education, Youth and Sports of the Czech Republic.

\section*{Author contributions statement}

    J.K. and K.H. proposed the time series metrics; J.K. designed and implemented the architecture for collecting time series metrics and maintained the collection pipeline over the extended multi-year period; K.H. and P.\v{S}. designed and implemented the architecture for capturing network traffic; K.J. maintained the flow-monitoring infrastructure and prepared the archived longitudinal-subnet data for re-aggregation; J.K. and K.H. preprocessed the data into the final version of the dataset; J.K. performed the long-term technical-validation analyses (annual seasonality, reproducibility, drift, model longevity, and calendar-anomaly separation); K.H. and K.J. handled supervision; and all authors reviewed the manuscript.

\section*{Competing interests}

    The authors declare no competing interests.

\bibliographystyle{unsrt}  
\bibliography{sample}

@article{koumar2025cesnet,
  title={Cesnet-timeseries24: Time series dataset for network traffic anomaly detection and forecasting},
  author={Koumar, Josef and Hynek, Karel and {\v{C}}ejka, Tom{\'a}{\v{s}} and {\v{S}}i{\v{s}}ka, Pavel},
  journal={Scientific Data},
  volume={12},
  number={1},
  pages={338},
  year={2025},
  publisher={Nature Publishing Group UK London}
}

@inproceedings{jerabek2025explainable,
  title={Explainable anomaly detection in network traffic using llm},
  author={Jerabek, Kamil and Koumar, Josef and Setinsk{\`y}, Ji{\v{r}}{\'\i} and Pesek, Jaroslav},
  booktitle={NOMS 2025-2025 IEEE Network Operations and Management Symposium},
  pages={1--6},
  year={2025},
  organization={IEEE}
}

@article{jerabek2025simple,
  title={When simple model just works: Is network traffic classification in crisis?},
  author={Jerabek, Kamil and Luxemburk, Jan and Plny, Richard and Koumar, Josef and Pesek, Jaroslav and Hynek, Karel},
  journal={arXiv preprint arXiv:2506.08655},
  year={2025}
}

@InProceedings{10.1007/978-3-642-30507-8_13,
author="Syarif, Iwan
and Prugel-Bennett, Adam
and Wills, Gary",
editor="Benlamri, Rachid",
title="Unsupervised Clustering Approach for Network Anomaly Detection",
booktitle="Networked Digital Technologies",
year="2012",
publisher="Springer Berlin Heidelberg",
address="Berlin, Heidelberg",
pages="135--145",
isbn="978-3-642-30507-8"
}

@ARTICLE{8990084,
  author={Hwang, Ren-Hung and Peng, Min-Chun and Huang, Chien-Wei and Lin, Po-Ching and Nguyen, Van-Linh},
  journal={IEEE Access}, 
  title={An Unsupervised Deep Learning Model for Early Network Traffic Anomaly Detection}, 
  year={2020},
  volume={8},
  number={},
  pages={30387-30399},
  doi={10.1109/ACCESS.2020.2973023}
}

@article{wright1921correlation,
  title={Correlation and causation},
  author={Wright, Sewall},
  journal={Journal of agricultural research},
  volume={20},
  number={7},
  pages={557},
  year={1921}
}

@article{chicco2021coefficient,
  title={The coefficient of determination R-squared is more informative than SMAPE, MAE, MAPE, MSE and RMSE in regression analysis evaluation},
  author={Chicco, Davide and Warrens, Matthijs J and Jurman, Giuseppe},
  journal={Peerj computer science},
  volume={7},
  pages={e623},
  year={2021},
  publisher={PeerJ Inc.}
}

@inproceedings{andrysiak2014network,
  title={Network traffic prediction and anomaly detection based on ARFIMA model},
  author={Andrysiak, Tomasz and Saganowski, {\L}ukasz and Chora{\'s}, Micha{\l} and Kozik, Rafa{\l}},
  booktitle={International Joint Conference SOCO’14-CISIS’14-ICEUTE’14: Bilbao, Spain, June 25th-27th, 2014, Proceedings},
  pages={545--554},
  year={2014},
  organization={Springer}
}

@inproceedings{yaacob2010arima,
  title={Arima based network anomaly detection},
  author={Yaacob, Asrul H and Tan, Ian KT and Chien, Su Fong and Tan, Hon Khi},
  booktitle={2010 Second International Conference on Communication Software and Networks},
  pages={205--209},
  year={2010},
  organization={IEEE}
}

@article{ferreira2023forecasting,
  title={Forecasting network traffic: A survey and tutorial with open-source comparative evaluation},
  author={Ferreira, Gabriel O and Ravazzi, Chiara and Dabbene, Fabrizio and Calafiore, Giuseppe C and Fiore, Marco},
  journal={IEEE Access},
  volume={11},
  pages={6018--6044},
  year={2023},
  publisher={IEEE}
}

@inproceedings{mawilab,
 author = {Fontugne, Romain and Borgnat, Pierre and Abry, Patrice and Fukuda, Kensuke},
 title = {{MAWILab: Combining Diverse Anomaly Detectors for Automated Anomaly Labeling and Performance Benchmarking}},
 booktitle = {ACM CoNEXT '10},
 month = {December},
 year = {2010},
 address = {Philadelphia, PA},
 numpages = {12}
}

@inproceedings{benes2023look,
  title={Look at my Network: An insight into the ISP Backbone Traffic},
  author={Benes, Tomas and Pesek, Jaroslav and Cejka, Tomas},
  booktitle={2023 19th International Conference on Network and Service Management (CNSM)},
  pages={1--7},
  year={2023},
  organization={IEEE}
}

@article{d2019survey,
  title={A survey on big data for network traffic monitoring and analysis},
  author={D’Alconzo, Alessandro and Drago, Idilio and Morichetta, Andrea and Mellia, Marco and Casas, Pedro},
  journal={IEEE Transactions on Network and Service Management},
  volume={16},
  number={3},
  pages={800--813},
  year={2019},
  publisher={IEEE}
}

@article{akbari2022traffic,
  title={Traffic classification in an increasingly encrypted web},
  author={Akbari, Iman and Salahuddin, Mohammad A and Ven, Leni and Limam, Noura and Boutaba, Raouf and Mathieu, Bertrand and Moteau, Stephanie and Tuffin, Stephane},
  journal={Communications of the ACM},
  volume={65},
  number={10},
  pages={75--83},
  year={2022},
  publisher={ACM New York, NY, USA}
}

@article{aceto2019mobile,
  title={Mobile encrypted traffic classification using deep learning: Experimental evaluation, lessons learned, and challenges},
  author={Aceto, Giuseppe and Ciuonzo, Domenico and Montieri, Antonio and Pescap{\'e}, Antonio},
  journal={IEEE transactions on network and service management},
  volume={16},
  number={2},
  pages={445--458},
  year={2019},
  publisher={IEEE}
}

@article{koumar2024nettisa,
  title={NetTiSA: Extended IP flow with time-series features for universal bandwidth-constrained high-speed network traffic classification},
  author={Koumar, Josef and Hynek, Karel and Pe{\v{s}}ek, Jaroslav and {\v{C}}ejka, Tom{\'a}{\v{s}}},
  journal={Computer Networks},
  volume={240},
  pages={110147},
  year={2024},
  publisher={Elsevier}
}

@inproceedings{plny2022decrypto,
  title={DeCrypto: Finding cryptocurrency miners on ISP networks},
  author={Pln{\`y}, Richard and Hynek, Karel and {\v{C}}ejka, Tom{\'a}{\v{s}}},
  booktitle={Nordic Conference on Secure IT Systems},
  pages={139--158},
  year={2022},
  organization={Springer}
}

@article{wu2021current,
  title={Current time series anomaly detection benchmarks are flawed and are creating the illusion of progress},
  author={Wu, Renjie and Keogh, Eamonn J},
  journal={IEEE transactons on knowledge and data engineering},
  volume={35},
  number={3},
  pages={2421--2429},
  year={2021},
  organization={IEEE}
}

@inproceedings{cho2000traffic,
  title={Traffic data repository at the $\{$WIDE$\}$ project},
  author={Cho, Kenjiro and Mitsuya, Koushirou and Kato, Akira},
  booktitle={2000 USENIX Annual Technical Conference (USENIX ATC 00)},
  year={2000}
}

@article{bhuyan2011surveying,
  title={Surveying port scans and their detection methodologies},
  author={Bhuyan, Monowar H and Bhattacharyya, Dhruba Kr and Kalita, Jugal K},
  journal={The Computer Journal},
  volume={54},
  number={10},
  pages={1565--1581},
  year={2011},
  publisher={Oxford University Press}
}

@article{staniford2002practical,
  title={Practical automated detection of stealthy portscans},
  author={Staniford, Stuart and Hoagland, James A and McAlerney, Joseph M},
  journal={Journal of Computer Security},
  volume={10},
  number={1-2},
  pages={105--136},
  year={2002},
  publisher={IOS Press}
}

@misc{rfc7011,
    series =    {Request for Comments},
    number =    7011,
    howpublished =  {RFC 7011},
    publisher = {RFC Editor},
    doi =       {10.17487/RFC7011},
    url =       {https://www.rfc-editor.org/info/rfc7011},
    author =    {Paul Aitken and Benoît Claise and Brian Trammell},
    title =     {{Specification of the IP Flow Information Export (IPFIX) Protocol for the Exchange of Flow Information}},
    pagetotal = 76,
    year =      2013,
    month =     sep,
}

@article{chang2026time,
  title={Time-imm: A dataset and benchmark for irregular multimodal multivariate time series},
  author={Chang, Ching and Hwang, Jeehyun and Shi, Yidan and Wang, Haixin and Wang, Wei and Peng, Wen-Chih and Chen, Tien-Fu},
  journal={Advances in Neural Information Processing Systems},
  volume={38},
  year={2026}
}

@article{liu2025time,
  title={Time-series foundation models for isp traffic forecasting},
  author={Liu, Fan and Farkiani, Behrooz and Crowley, Patrick},
  journal={arXiv preprint arXiv:2511.17529},
  year={2025}
}

@article{koumar2025comparative,
  title={Comparative analysis of deep learning models for real-world isp network traffic forecasting},
  author={Koumar, Josef and Smole{\v{n}}, Timotej and Je{\v{r}}{\'a}bek, Kamil and {\v{C}}ejka, Tom{\'a}{\v{s}}},
  journal={IEEE Transactions on Network and Service Management},
  volume={23},
  pages={715--728},
  year={2025},
  publisher={IEEE}
}

@article{feng2026robust,
  title={Robust long-term spatial--temporal forecasting for dynamic networks: The mdetst model},
  author={Feng, Wenjing and Li, Lanhao and Zhou, Huichao and Xie, Bingshu and Zhao, Dawei and Zhang, Yi and Wang, Feifei},
  journal={Mathematics},
  volume={14},
  number={4},
  pages={661},
  year={2026},
  publisher={MDPI}
}

@article{sengendo2026prediction,
  title={From Prediction to Protection: A Network Digital Twin framework for Traffic Monitoring and Anomaly Detection},
  author={Sengendo, John and Granelli, Fabrizio},
  journal={Computer Networks},
  pages={112250},
  year={2026},
  publisher={Elsevier}
}

@inproceedings{misa2025sustainability,
  title={Sustainability or Survivability? Eliminating the Need to Choose in LEO Satellite Constellations},
  author={Misa, Chris and Durairajan, Ramakrishnan},
  booktitle={Proceedings of the 24th ACM Workshop on Hot Topics in Networks},
  pages={123--130},
  year={2025}
}

@article{zhang2026centile,
  title={CENTILE: A Telemetry Foundation Model Evaluated by the Decisions It Drives},
  author={Zhang, Zifan and Hou, Zhichao and Ji, Tingxiang and Liu, Yuchen},
  journal={arXiv preprint arXiv:2608.01725},
  year={2026}
}

@article{babai2026deep,
  title={Which Deep Learner? A Systematic Evaluation of Advanced Deep Forecasting Models Accuracy and Efficiency for Network Traffic Prediction},
  author={Babai, Eilaf and Babai, Aalaa and Okamura, Koji},
  journal={arXiv preprint arXiv:2601.02694},
  year={2026}
}

@inproceedings{10.1145/781027.781053,
author = {Zhang, Yin and Roughan, Matthew and Duffield, Nick and Greenberg, Albert},
title = {Fast accurate computation of large-scale IP traffic matrices from link loads},
year = {2003},
isbn = {1581136641},
publisher = {Association for Computing Machinery},
address = {New York, NY, USA},
url = {https://doi.org/10.1145/781027.781053},
doi = {10.1145/781027.781053},
booktitle = {Proceedings of the 2003 ACM SIGMETRICS International Conference on Measurement and Modeling of Computer Systems},
pages = {206–217},
numpages = {12},
location = {San Diego, CA, USA},
series = {SIGMETRICS '03}
}

@article{10.1145/1111322.1111341,
author = {Uhlig, Steve and Quoitin, Bruno and Lepropre, Jean and Balon, Simon},
title = {Providing public intradomain traffic matrices to the research community},
year = {2006},
issue_date = {January 2006},
publisher = {Association for Computing Machinery},
address = {New York, NY, USA},
volume = {36},
number = {1},
issn = {0146-4833},
url = {https://doi.org/10.1145/1111322.1111341},
doi = {10.1145/1111322.1111341},
journal = {SIGCOMM Comput. Commun. Rev.},
month = jan,
pages = {83–86},
numpages = {4}
}

@INPROCEEDINGS{9500331,
  author={Le, Duc-Huy and Tran, Hai-Anh and Souihi, Sami and Mellouk, Abdelhamid},
  booktitle={ICC 2021 - IEEE International Conference on Communications}, 
  title={An AI-based Traffic Matrix Prediction Solution for Software-Defined Network}, 
  year={2021},
  volume={},
  number={},
  pages={1-6},
  doi={10.1109/ICC42927.2021.9500331}
}

@article{barlacchi2015multi,
  title={A multi-source dataset of urban life in the city of Milan and the Province of Trentino},
  author={Barlacchi, Gianni and De Nadai, Marco and Larcher, Roberto and Casella, Antonio and Chitic, Cristiana and Torrisi, Giovanni and Antonelli, Fabrizio and Vespignani, Alessandro and Pentland, Alex and Lepri, Bruno},
  journal={Scientific data},
  volume={2},
  number={1},
  pages={150055},
  year={2015},
  publisher={Nature Publishing Group}
}

@dataset{fondo_ferreiro_2023_8147768,
  author       = {Fondo-Ferreiro, Pablo and
                  Rodelgo-Lacruz, Miguel and
                  González-Castaño, Francisco Javier and
                  Gil-Castiñeira, Felipe and
                  Candal-Ventureira, David},
  title        = {Network operator KPIs time series dataset},
  month        = jul,
  year         = 2023,
  publisher    = {Zenodo},
  version      = {1.0},
  doi          = {10.5281/zenodo.8147768},
  url          = {https://doi.org/10.5281/zenodo.8147768},
}

@article{orlowski2010sndlib,
  title={SNDlib 1.0—Survivable network design library},
  author={Orlowski, Sebastian and Wess{\"a}ly, Roland and Pi{\'o}ro, Michal and Tomaszewski, Artur},
  journal={Networks: An International Journal},
  volume={55},
  number={3},
  pages={276--286},
  year={2010},
  publisher={Wiley Online Library}
}

@article{wang2006characteristic,
  title={Characteristic-Based Clustering for Time Series Data: Wang, Smith and Hyndman},
  author={Wang, Xiaozhe and Smith, Kate and Hyndman, Rob},
  journal={Data mining and knowledge Discovery},
  volume={13},
  number={3},
  pages={335--364},
  year={2006},
  publisher={Springer}
}

@dataset{CESNET_TIMESERIES26_DATASET,
  author       = {Koumar, Josef and
                  Hynek, Karel and
                  Jeřábek, Kamil and
                  Šiška, Pavel},
  title        = {CESNET-TimeSeries26: A Multi-Year Network Traffic Time Series Dataset for Annual Seasonality, Concept Drift, and Long-Term Anomaly Detection},
  month        = sep,
  year         = 2026,
  publisher    = {Zenodo},
  doi          = {10.5281/zenodo.21945629},
  url          = {https://doi.org/10.5281/zenodo.21945629},
}

@article{liu2023largest,
  title={Largest: A benchmark dataset for large-scale traffic forecasting},
  author={Liu, Xu and Xia, Yutong and Liang, Yuxuan and Hu, Junfeng and Wang, Yiwei and Bai, Lei and Huang, Chao and Liu, Zhenguang and Hooi, Bryan and Zimmermann, Roger},
  journal={Advances in neural information processing systems},
  volume={36},
  pages={75354--75371},
  year={2023}
}

@article{makridakis2022m5,
  title={M5 accuracy competition: Results, findings, and conclusions},
  author={Makridakis, Spyros and Spiliotis, Evangelos and Assimakopoulos, Vassilios},
  journal={International journal of forecasting},
  volume={38},
  number={4},
  pages={1346--1364},
  year={2022},
  publisher={Elsevier}
}

@article{godahewa2021monash,
  title={Monash time series forecasting archive},
  author={Godahewa, Rakshitha and Bergmeir, Christoph and Webb, Geoffrey I and Hyndman, Rob J and Montero-Manso, Pablo},
  journal={arXiv preprint arXiv:2105.06643},
  year={2021}
}

@inproceedings{lai2018modeling,
  title={Modeling long-and short-term temporal patterns with deep neural networks},
  author={Lai, Guokun and Chang, Wei-Cheng and Yang, Yiming and Liu, Hanxiao},
  booktitle={The 41st international ACM SIGIR conference on research \& development in information retrieval},
  pages={95--104},
  year={2018}
}

@inproceedings{zhou2021informer,
  title={Informer: Beyond efficient transformer for long sequence time-series forecasting},
  author={Zhou, Haoyi and Zhang, Shanghang and Peng, Jieqi and Zhang, Shuai and Li, Jianxin and Xiong, Hui and Zhang, Wancai},
  booktitle={Proceedings of the AAAI conference on artificial intelligence},
  volume={35},
  number={12},
  pages={11106--11115},
  year={2021}
}

@article{wu2021autoformer,
  title={Autoformer: Decomposition transformers with auto-correlation for long-term series forecasting},
  author={Wu, Haixu and Xu, Jiehui and Wang, Jianmin and Long, Mingsheng},
  journal={Advances in neural information processing systems},
  volume={34},
  pages={22419--22430},
  year={2021}
}

@inproceedings{kurevs2025cesnet,
  title={CESNET TS-Zoo: A Library for Reproducible Analysis of Network Traffic Time Series},
  author={Kure{\v{s}}, Milan and Koumar, Josef and Hynek, Karel},
  booktitle={2025 21st International Conference on Network and Service Management (CNSM)},
  pages={1--5},
  year={2025},
  organization={IEEE}
}

\appendix

\begin{table*}
          \centering
          \caption{\label{tab:aggregation_metrics}Detailed descriptions of time series metrics for each IP address dataset. Note: in the longitudinal subnet part, all volumetric metrics reflect TCP traffic only, and \texttt{tcp\_udp\_ratio\_packets}/\texttt{tcp\_udp\_ratio\_bytes} are constant ($=1$); see the Longitudinal subnet collection subsection.}
          \rowcolors{3}{gray!10}{white}
          \begin{tabular}{m{2.5cm}|m{14cm}}
              \toprule
              \textbf{Column name } & \textbf{Description} \\
              \midrule
              \texttt{id\_time} & Unique identifier for each aggregation interval within the time series, used to segment the dataset into specific time periods for analysis. \\
              \texttt{n\_flows} & Total number of flows observed in the aggregation interval, indicating the volume of distinct sessions or connections for the IP address. \\
              \texttt{n\_packets} & Total number of packets transmitted during the aggregation interval, reflecting the packet-level traffic volume for the IP address. \\
              \texttt{n\_bytes} & Total number of bytes transmitted during the aggregation interval, representing the data volume for the IP address. \\
              \texttt{n\_dest\_ip} & Number of unique destination IP addresses contacted by the IP address during the aggregation interval, showing the diversity of endpoints reached. \\
              \texttt{n\_dest\_asn} & Number of unique destination Autonomous System Numbers (ASNs) contacted by the IP address during the aggregation interval, indicating the diversity of networks reached. \\
              \texttt{n\_dest\_port} & Number of unique destination transport layer ports contacted by the IP address during the aggregation interval, representing the variety of services accessed. \\
              \texttt{tcp\_udp\_}\allowbreak\texttt{ratio\_}\allowbreak\texttt{packets} & Ratio of packets sent using TCP versus UDP by the IP address during the aggregation interval, providing insight into the transport protocol usage pattern. This metric belongs to the interval $\langle 0, 1\rangle$ where 1 is when all packets are sent over TCP, and 0 is when all packets are sent over UDP. \\
              \texttt{tcp\_udp\_}\allowbreak\texttt{ratio\_}\allowbreak\texttt{bytes} & Ratio of bytes sent using TCP versus UDP by the IP address during the aggregation interval, highlighting the data volume distribution between protocols. This metric belongs to the interval $\langle 0, 1\rangle$ with the same rule as \texttt{tcp\_udp\_ratio\_packets}. \\
              \texttt{dir\_}\allowbreak\texttt{ratio\_}\allowbreak\texttt{packets} & Ratio of packet directions (inbound versus outbound) for the IP address during the aggregation interval, indicating the balance of traffic flow directions. This metric belongs to the interval $\langle 0, 1\rangle$, where 1 is when all packets are sent in the outgoing direction from the monitored IP address, and 0 is when all packets are sent in the incoming direction to the monitored IP address. \\
              \texttt{dir\_}\allowbreak\texttt{ratio\_}\allowbreak\texttt{bytes} & Ratio of byte directions (inbound versus outbound) for the IP address during the aggregation interval, showing the data volume distribution in traffic flows. This metric belongs to the interval $\langle 0, 1\rangle$ with the same rule as \texttt{dir\_ratio\_packets}. \\
              \texttt{avg\_duration} & Average duration of IP flows for the IP address during the aggregation interval, measuring the typical session length. \\
              \texttt{avg\_ttl} & Average Time To Live (TTL) of IP flows for the IP address during the aggregation interval, providing insight into the lifespan of packets. \\
              \bottomrule
          \end{tabular}
        \end{table*}

        \begin{table*}
          \centering
          \caption{\label{tab:aggregation_metrics_2}Time series metrics which replace metrics \texttt{n\_dest\_ip}, \texttt{n\_dest\_asn} and \texttt{n\_dest\_port} in the 1-hour and 1-day aggregations.}
          \rowcolors{3}{gray!10}{white}
          \begin{tabular}{p{4cm}|p{12cm}}
              \toprule
              \textbf{Column name } & \textbf{Description} \\
              \midrule
              \texttt{sum\_n\_dest\_ip} & Sum of numbers of unique destination IP addresses. \\
              \texttt{avg\_n\_dest\_ip} & The average number of unique destination IP addresses. \\
              \texttt{std\_n\_dest\_ip} & Standard deviation of numbers of unique destination IP addresses. \\
              \texttt{sum\_n\_dest\_asn} & Sum of numbers of unique destination ASNs. \\
              \texttt{avg\_n\_dest\_asn} & The average number of unique destination ASNs. \\
              \texttt{std\_n\_dest\_asn} & Standard deviation of numbers of unique destination ASNs. \\
              \texttt{sum\_n\_dest\_port} & Sum of numbers of unique destination transport layer ports. \\
              \texttt{avg\_n\_dest\_port} & The average number of unique destination transport layer ports. \\
              \texttt{std\_n\_dest\_port} & Standard deviation of numbers of unique destination transport layer ports. \\
              \bottomrule
          \end{tabular}

        \end{table*}

        \begin{table*}

          \centering
          \caption{\label{tab:id_relationship}Content of the \texttt{ids\_relationship.csv} file.}
          \rowcolors{3}{gray!10}{white}
          \begin{tabular}{p{4.6cm}|p{11.4cm}}
              \toprule
              \textbf{Column name } & \textbf{Description} \\
              \midrule
              \texttt{id\_ip} & ID of the IP address.  \\
              \texttt{id\_institution} & ID of the institution which owns the IP address. \\
              \texttt{id\_institution\_}\allowbreak\texttt{subnet} & ID of the institution subnet in which the IP address belongs. \\
              \bottomrule
          \end{tabular}

        \end{table*}

        \begin{table*}

          \centering
          \caption{\label{tab:weekends_and_holidays}Content of the \texttt{weekends\_and\_holidays.csv} file.}
          \rowcolors{3}{gray!10}{white}
          \begin{tabular}{p{4cm}|p{12cm}}
              \toprule
              \textbf{Column name } & \textbf{Description} \\
              \midrule
              \texttt{Date} & The date of the day in format ``YYYY-MM-DD''. \\
              \texttt{Type} & The type of the day (Weekend or Holiday). \\
              \bottomrule
          \end{tabular}

        \end{table*}

        \begin{table*}

        \centering
        \caption{\label{tab:school_terms}Content of the \texttt{school\_terms\_and\_exams\_periods.csv} file.}
        \rowcolors{3}{gray!10}{white}
        \begin{tabular}{p{4cm}|p{12cm}}
            \toprule
            \textbf{Column name } & \textbf{Description} \\
            \midrule
            \texttt{Start} & The first day of the period in format ``YYYY-MM-DD''. \\
            \texttt{End} & The last day of the period (inclusive) in format ``YYYY-MM-DD''. \\
            \texttt{Type} & The type of the academic period (\texttt{WinterTerm}, \texttt{WinterExams}, \texttt{SummerTerm} or
  \texttt{SummerExams}). \\
            \bottomrule
        \end{tabular}

        \end{table*}

        \begin{table*}

        \centering
        \caption{\label{tab:europe_time_changes}Content of the \texttt{europe\_time\_changes.csv} file.}
        \rowcolors{3}{gray!10}{white}
        \begin{tabular}{p{4cm}|p{12cm}}
            \toprule
            \textbf{Column name } & \textbf{Description} \\
            \midrule
            \texttt{date} & The date of the daylight-saving transition in format ``YYYY-MM-DD''. \\
            \texttt{transition} & The direction of the transition (\texttt{spring\_forward} or \texttt{fall\_back}). \\
            \texttt{measured\_from} & The dataset part in which the transition falls inside the collection span (\texttt{main} or
  \texttt{subnet\_longitudinal}). \\
            \texttt{local\_clock\_change} & The local wall-clock jump, e.g.\ \texttt{02:00 CET $\rightarrow$ 03:00 CEST}. \\
            \texttt{utc\_offset\_before} & The UTC offset of local time before the transition (\texttt{+01:00} for CET,
  \texttt{+02:00} for CEST). \\
            \texttt{utc\_offset\_after} & The UTC offset of local time after the transition. \\
            \texttt{offset\_change\_h} & The signed change of the UTC offset in hours ($+1$ for spring-forward, $-1$ for
  fall-back). \\
            \bottomrule
        \end{tabular}

        \end{table*}

        \begin{table*}

        \centering
        \caption{\label{tab:outages_columns}Content of the \texttt{outages.csv} file.}
        \rowcolors{3}{gray!10}{white}
        \begin{tabular}{p{4cm}|p{12cm}}
            \toprule
            \textbf{Column name } & \textbf{Description} \\
            \midrule
            \texttt{id} & Sequential ID of the outage. \\
            \texttt{part} & The dataset part affected by the outage (\texttt{main} or \texttt{subnet\_longitudinal}). \\
            \texttt{scope} & The extent of the outage (\texttt{infrastructure-wide} or \texttt{subnet-local}). \\
            \texttt{kind} & How the outage manifests in the data: \texttt{grid-gap} for windows absent from the shared time grid,
  \texttt{present-zero} for windows present in the grid but carrying zero traffic. \\
            \texttt{last\_data\_before} & Timestamp of the last window carrying data before the outage, with explicit UTC offset.
  \\
            \texttt{data\_resumed} & Timestamp of the first window carrying data after the outage, with explicit UTC offset. \\
            \texttt{duration} & The length of the outage in human-readable form, e.g.\ \texttt{15 d 21.0 h}. \\
            \texttt{n\_windows\_10min} & The number of 10-minute windows the outage spans. \\
            \bottomrule
        \end{tabular}

        \end{table*}

\end{document}